%% file: main.tex
\documentclass[11pt]{article}

\usepackage[a4paper,margin=2.5cm]{geometry}
\usepackage[T1]{fontenc}
\usepackage[utf8]{inputenc}
\usepackage{microtype}
\usepackage{amsmath,amssymb,mathtools,bm}
\usepackage{graphicx}
\usepackage{float}
\usepackage[section]{placeins}
\usepackage{booktabs}
\usepackage{siunitx}
\usepackage{xcolor}
\usepackage{authblk}
\usepackage{lineno}
\usepackage[numbers,sort&compress]{natbib}
\usepackage[colorlinks=true,allcolors=blue]{hyperref}
\usepackage[nameinlink,capitalise]{cleveref}

\graphicspath{{figures/}}
\nolinenumbers

\newcommand{\R}{\mathbb{R}}
\newcommand{\domain}{\Omega}
\newcommand{\refdomain}{\widehat{\Omega}}
\newcommand{\map}{\Phi}
\newcommand{\jac}{\bm{J}}
\newcommand{\keff}{\bm{K}}
\newcommand{\cofj}{\operatorname{cof}\jac}

\newcommand{\maybeincludegraphics}[2][]{%
  \IfFileExists{#2}{\includegraphics[#1]{#2}}{%
    \fbox{\parbox[c][45mm][c]{0.92\linewidth}{\centering
    Figure placeholder\\[0.5em]\texttt{\detokenize{#2}}}}}}

\title{Geometry--physics confounding impairs PDE learning across varying domains}

\author[1,2]{Yinghao Cheng}
\author[1,3]{Gengxiang Chen}
\author[4]{Xu Liu}
\author[1]{Qinglu Meng}
\author[1]{Yixin Jing}
\author[1]{Xiangguo Tang}
\author[1]{Wenping Mou}
\author[5,6]{Lihui Wang}
\author[1,*]{Yingguang Li}

\affil[1]{College of Mechanical \& Electrical Engineering, Nanjing University of Aeronautics and Astronautics, Nanjing, China}
\affil[2]{School of Electronics, Electrical Engineering and Computer Science, Queen's University Belfast, Belfast, United Kingdom}
\affil[3]{School of Mechanical and Aerospace Engineering, Queen's University Belfast, Belfast, United Kingdom}
\affil[4]{School of Mechanical and Power Engineering, Nanjing Tech University, Nanjing, China}
\affil[5]{Department of Industrial and Systems Engineering, The Hong Kong Polytechnic University, Hong Kong, China}
\affil[6]{Department of Production Engineering, KTH Royal Institute of Technology, Stockholm, Sweden}
\affil[*]{Correspondence: liyingguang@nuaa.edu.cn.}

\date{}

\begin{document}
\maketitle

\input{sections/00_abstract}

\input{sections/01_introduction}

\section*{Geometry--physics confounding and de-confounding methods}
\input{sections/02_framework}

\input{sections/results/03_01_neural_operator}
\input{sections/results/03_02_mechanism_identification}

\section*{Discussion}
\input{sections/05_discussion}

\section*{Methods}
\input{sections/methods/04_01_geometry_and_diagnostic}
\input{sections/methods/04_02_operator_learning}
\input{sections/methods/04_03_discovery_and_evaluation}

\section*{Competing interests}
The authors declare no competing interests.

\section*{Additional information}
Supplementary information accompanies this manuscript.

\bibliographystyle{unsrt}
\bibliography{references}

\clearpage
\input{sections/06_supplementary}

\end{document}

%% file: sections/00_abstract.tex
\begin{abstract}
\noindent
Learning partial differential equation (PDE) dynamics across varying domains is central to predictive modelling and data-driven discovery of governing equations. However, geometric variation alters both field representation and the governing differential operators, confounding geometric effects with intrinsic physical properties in the observed dynamics. This work identifies \emph{geometry--physics confounding} as a unified failure mechanism for PDE learning across varying domains. In forward operator learning, this confounding increases the burden of inferring geometry-dependent operator changes from finite data, reducing data efficiency and generalisation. In equation discovery, omitting geometry-induced operators misspecifies the candidate library, leading to biased parameters, missed governing terms and spurious terms. We propose a de-confounding framework that makes the known geometry-to-operator transformation explicit. Geometry-induced coefficient fields improve prediction and data efficiency across five operator-learning benchmarks, while geometry-complete candidate libraries recover the generating equations and reduce held-out PDE residuals by more than two orders of magnitude in both evolving-domain systems. By separating known geometric action from intrinsic physics, the proposed framework supports more reliable and data-efficient PDE learning across scientific and engineering problems with varying geometries.
\end{abstract}

%% file: sections/01_introduction.tex
\noindent Partial differential equations (PDEs) on varying and evolving domains govern a broad spectrum of transport, wave, and deformation phenomena across continuum physics and biology, from turbulent aerodynamics over complex geometries \cite{vinuesa2023transformative,li2023geofno,wu2024transolver} and patient-specific cardiac mechanics and forward electrocardiographic modelling \cite{trayanova2011wholeheart,prakosa2018virtualheart,carrara2026shape,dokuchaev2026leadfield} to finite-strain elastodynamics \cite{niederer2021digitaltwins}, morphogenetic tissue growth \cite{friedman2015freeboundary}, and moving phase boundaries \cite{chen2008phasefield,dziuk2013surfacepdes,long2026freeboundary}. Learning PDE dynamics across varying geometries from spatiotemporal observations, encompassing solution-operator learning for predictive simulation \cite{lu2021deeponet,li2021fno,kovachki2023neuraloperator,azizzadenesheli2024neural,yin2024dimon} and data-driven discovery of governing equations \cite{brunton2016sindy,rudy2017pdefind,rao2023encoding,course2023state,messenger2021wsindy}, is an increasingly important computational foundation for scientific discovery and engineering design. Since geometric variation changes not only the domain on which the state is represented but also the differential operators governing its evolution, capturing how geometry transforms these operators is a central challenge for PDE learning across varying domains.

Recent advances in neural operators have enabled the learning of PDE solution operators between function spaces \cite{lu2021deeponet,li2021fno,kovachki2023neuraloperator}. For PDEs on complex geometries, early extensions typically mapped irregular physical domains to regular computational domains, using learned deformations to latent grids in Geo-FNO \cite{li2023geofno}, graph-based lifting to Cartesian latent grids in GINO \cite{li2023gino}, or diffeomorphic mappings to a common reference domain in DIMON \cite{yin2024dimon}. As operator learning has been extended to increasingly complex and unstructured geometries, recent methods have incorporated geometric information directly into the model through coordinates, point clouds, meshes, or continuous geometric representations, as exemplified by neural-field operators \cite{serrano2023coral}, physics-aware attention in Transolver \cite{wu2024transolver}, and more recent graph- and transformer-based architectures for large-scale irregular domains \cite{luo2025transolverpp,mousavi2025rigno,wen2025gaot}. These developments have substantially expanded the applicability of neural operators to PDEs on varying geometries. Their treatment of geometry remains primarily concerned with the representation and processing of the computational domain, without explicitly accounting for how geometric variation alters the governing differential operators. Consequently, these geometry-dependent operator changes must still be inferred implicitly from finite training data.

Learning PDE dynamics from observations collected across varying domains requires correspondence between fields on different geometries, established explicitly through reference maps or implicitly through geometry-conditioned representations. A reference map places these fields on a common domain, but the same pull-back also transforms the spatial differential operators and, for moving domains, introduces transport associated with domain motion \cite{hughes1981ale,donea2004ale,dziuk2013surfacepdes}. Domain-mapping theory establishes the equivalence between varying-domain PDEs and their transformed problems on a fixed reference domain \cite{harbrecht2016domainmapping}, while recent theory for neural shape operators provides well-posedness, shape regularity and approximation-rate results for the corresponding shape-to-solution maps \cite{harbrecht2026shape}. Related studies have shown that operator-preserving conformal mappings can improve the learning of Laplace solution operators \cite{ahmad2024diffeomorphic}, while boundary-adapted coordinates can support the discovery of moving-boundary dynamics \cite{bekar2025multiphysics}. These task-specific treatments demonstrate the value of respecting geometric structure, but leave unresolved whether a learned or discovered model can distinguish known geometric action from intrinsic physical relations when both enter the transformed dynamics through the same coefficients.

We refer to this non-unique physical attribution as \emph{geometry--physics confounding}. When a PDE is expressed on a common reference domain, geometric transformation changes its differential operators: homogeneous and isotropic physics can appear through spatially varying and anisotropic coefficients, while time-dependent mappings additionally introduce arbitrary Lagrangian--Eulerian (ALE) transport associated with domain motion \cite{harbrecht2016domainmapping,hughes1981ale,donea2004ale,dziuk2013surfacepdes,bekar2025multiphysics}. Physically distinct combinations of geometry and intrinsic properties can therefore generate identical reference-domain dynamics and cannot be attributed uniquely from those dynamics alone. This ambiguity is conditional on the information used for attribution: supplying the domain map makes the decomposition available, but only if the learning or discovery procedure represents its PDE-specific action. In operator learning, leaving the geometric action implicit adds the finite-data burden of learning geometry-dependent operator changes from examples. In equation discovery, a transformed operator that is absent from the candidate library cannot be selected, leading to missed governing terms and the absorption of geometric effects into biased coefficients or spurious terms.

Here we establish geometry--physics confounding as a unified failure mechanism in operator prediction and equation discovery across varying domains. A diffusion example demonstrates exact observational equivalence and shows how implicit geometric action increases the finite-data learning burden in prediction and causes structural candidate-library misspecification in equation discovery. To address these failures, we propose a de-confounding framework that supplies geometry-induced coefficient fields for operator learning and geometry-complete candidate libraries for equation discovery. The framework modifies the problem representation within existing neural-operator formulations and retains the sparse-regression algorithm. Five operator-learning benchmarks spanning reference-domain mapping and geometry-aware models evaluate prediction and data efficiency \cite{yin2024dimon,li2023geofno,wu2024transolver}, while thermoelastic and free-boundary systems evaluate equation discovery and closed-loop transfer on evolving domains. Across these experiments, making geometric action explicit improves finite-data prediction across all five operator-learning benchmarks and enables recovery of the generating equations in both evolving-domain systems.

%% file: sections/02_framework.tex
\subsection*{Problem setup}

\begin{figure}[p]
  \centering
  \maybeincludegraphics[width=\linewidth]{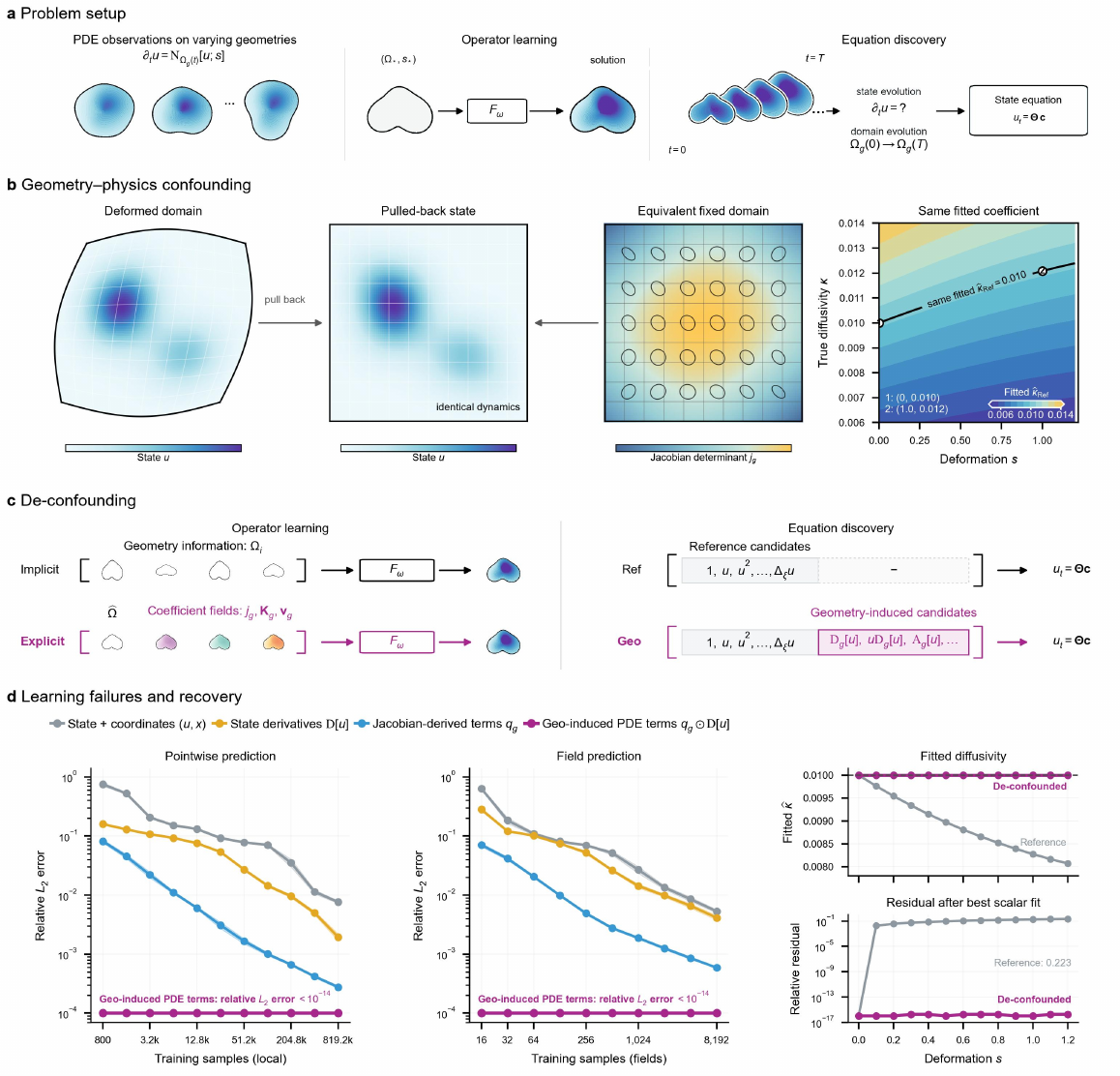}
  \caption{\textbf{Geometry--physics confounding in PDE learning.}
  \textbf{a}, Problem setup for PDE learning on varying geometries, illustrating sample observations pairing geometry with field data, operator learning of solution maps $F_\omega$, and equation discovery identifying governing differential laws.
  \textbf{b}, Geometry--physics confounding mechanism: a homogeneous isotropic diffusion process on a deformed domain and an equivalent fixed-domain model with geometry-induced capacity $j_g$ and transformed conductivity $\widehat{\bm\kappa}_g$ yield identical pulled-back dynamics. Fitting a misspecified reference model yields identical apparent diffusion coefficients ($\widehat\kappa_{\rm Ref}=0.010$) across physically distinct systems (contour plot, right).
  \textbf{c}, De-confounding strategies: geometry-induced coefficient fields for operator learning (left) and geometry-complete candidate libraries for equation discovery (right).
  \textbf{d}, Learning failures and recovery: for operator rate prediction (left and centre), coordinate-based inputs yield high errors, while explicitly incorporating geometry-induced coefficient fields reduces error by over fivefold, and the geometry-induced PDE terms yield numerical-precision predictions ($<10^{-14}$). Prediction curves and shaded bands show means and $\pm1$~s.d.\ across five optimisation runs. For equation discovery (right), the apparent reference coefficient exhibits a 19.3\% geometric drift and an equation residual of $0.223$, whereas the geometry-complete candidate library recovers the generating coefficient with numerical-precision residual ($<10^{-16}$).}
  \label{fig:confounding}
\end{figure}

As shown in Fig.~\ref{fig:confounding}a, we consider PDE-learning samples
associated with geometries $g\in\mathcal G$. For each geometry, we consider a physical domain
$\domain_g(t)\subset\R^d$, a state field $u$, and problem data $s$ comprising
initial and boundary conditions, source terms, PDE coefficients and control inputs.
We represent the PDE problem compactly as follows, with the initial
and boundary conditions specified in $s$ understood implicitly:
\begin{equation}
 \partial_t u=\mathcal N_{\domain_g(t)}
 [u;s],
 \qquad x\in\domain_g(t),\quad t\in(0,T],
 \label{eq:varying-domain-problem}
\end{equation}
where $\partial_t$ denotes the time derivative associated with the
physical description, and $\mathcal N$ denotes the remaining physical-domain
differential operator. Steady problems satisfy
$0=\mathcal N_{\domain_g}[u;s]$. 

In real-world
applications, the geometries may represent a family of
related objects under study, such as corresponding blood vessels or organs
across different patients or alternative airfoil shapes explored in aircraft
design. They may also evolve over time through a moving-boundary process, as in
glacier melt or tumour growth. When domain evolution is coupled to the state, a mechanical 
or kinematic evolution law is required to close the coupled state--geometry problem.

Assuming that the physical problem is well posed for the prescribed data and
geometry, operator learning approximates the
geometry-conditioned solution map
\begin{equation}
 \mathcal F_{\bm\omega}:
 \bigl(s,\domain_g(\cdot)\bigr)\longmapsto u,
 \label{eq:varying-domain-operator-learning}
\end{equation}
where one parameter vector $\bm\omega$ is shared across the sampled geometries, and
the objective is to predict solution fields or trajectories for previously unseen prescribed problem data, 
geometries, or their combinations.

Equation discovery, by contrast, starts from observations across geometries and
a prespecified library of $p$ candidate operators defined on the physical domain,
and seeks
\begin{equation}
 \bigl(u,s^{\rm known},\domain_g(\cdot)\bigr)
 \longmapsto
 \bm c^\star,
 \label{eq:varying-domain-discovery}
\end{equation}
where $s^{\rm known}$ denotes the part of $s$ actually supplied during
identification, excluding the candidate coefficients to be inferred. The vector
$\bm c^\star\in\R^p$ contains the coefficients of the full candidate library,
with nonzero entries identifying the active terms. The
coefficient vector $\bm c^\star$ is shared across the sampled geometries.
Although the two tasks seek different objects, both concern PDEs across related
geometries or on evolving domains, with the same geometric transformations of
the governing equations.

The vessel, organ and airfoil examples above, together with successive domains
in tumour growth, naturally form related geometry families. We consider
topology-preserving families for which each physical domain is related to a
fixed reference domain $\refdomain\subset\R^d$ by a sufficiently regular,
orientation-preserving diffeomorphism
\begin{equation}
 \map_g(\cdot,t):\refdomain\to\domain_g(t),\qquad
 \jac_g=\nabla_\xi\map_g,\qquad
 j_g=\det\jac_g>0,
 \label{eq:reference-map}
\end{equation}
with time dependence suppressed for a time-independent map.
Under the regularity assumptions required by the chosen formulation, $\map_g$ induces
a pullback from physical-domain fields to reference-domain fields
\cite{harbrecht2016domainmapping,harbrecht2026shape,gong2026shapedino}
\begin{equation}
 \widehat u=\mathcal P_g u,\qquad
 (\mathcal P_g u)(\xi,t):=u\bigl(\map_g(\xi,t),t\bigr),\qquad
 \widehat s=\mathcal P_g^s s,
 \label{eq:reference-correspondence}
\end{equation}
where $\mathcal P_g$ represents the same state field on the reference domain by
evaluating it at corresponding physical points, and $\mathcal P_g^s$ denotes the corresponding transformations of
the prescribed problem data. Applying the chain rule and changes of variables
to Eq.~\eqref{eq:varying-domain-problem} yields the reference-domain evolution
equation
\begin{equation}
 \left.\partial_t\widehat u\right|_\xi
 =\widehat{\mathcal N}_{g,t}
 [\widehat u;\widehat s],
 \qquad \xi\in\refdomain,\quad t\in(0,T],
 \label{eq:pulled-back-evolution}
\end{equation}
where $\widehat{\mathcal N}_{g,t}$ denotes the reference-domain evolution
operator obtained by transforming Eq.~\eqref{eq:varying-domain-problem},
including its time derivative. Under the stated
regularity assumptions, equivalence follows directly from the invertibility of
$\map_g$. Specifically, $\mathcal P_g$ and its inverse establish a one-to-one correspondence
between solutions of the consistently transformed problems. The two
formulations describe the same physical evolution in different coordinates.
Although $\refdomain$ is fixed, the transformed operator retains geometric
dependence through the spatial transformation and any map-motion contributions
required by the adopted time description. Thus, the same physical-domain law
admits a family of reference-domain formulations in which geometry enters
through the transformed operators and problem data. We next examine how an
incomplete representation of geometric transformations can obscure physical
attribution and create distinct difficulties for operator learning and equation
discovery.

\subsection*{Geometry--physics confounding and failure mechanism}

To examine this attribution problem, consider diffusion on $\domain_g$ with a
time-independent map:
\begin{equation}
 \partial_tu=\kappa\Delta_xu,
 \qquad x\in\domain_g,
 \label{eq:physical-diffusion}
\end{equation}
where $\kappa$ denotes the physical-domain diffusion coefficient. Pulling this
equation back through the time-independent map to $\refdomain$ gives
\begin{equation}
 \partial_t \widehat{u}=
 \frac{1}{j_g}\nabla_{\xi}\!\cdot\!
 \left(\widehat{\bm\kappa}_g\nabla_{\xi}\widehat{u}\right),
 \qquad
 \widehat{\bm\kappa}_g
 =\kappa j_g\jac_g^{-1}\jac_g^{-\mathsf T}.
 \label{eq:pullback-diffusion}
\end{equation}
Thus homogeneous isotropic diffusion in physical coordinates appears on the
reference domain with the Jacobian determinant $j_g$ and transformed diffusion
coefficient field $\widehat{\bm\kappa}_g$. In the equivalent mass or weak form,
$j_g$ acts as a spatially varying capacity. The same spatial
transformation applies at each time for an evolving geometry, while the time-dependent map additionally
introduces transport induced by domain motion (Methods).
For operator learning, $\widehat{\bm\kappa}_g$ forms part of the equivalent
reference-domain problem data and is supplied as a complete field. In the
discovery experiment below, candidate coefficients are assigned to the
transformed operators and inferred from data.

Equation~\eqref{eq:pullback-diffusion} shows how two physically distinct
descriptions can produce the same pulled-back state trajectory when attribution
does not use the domain map separately. A homogeneous material on a
deformed domain and a heterogeneous anisotropic material on a fixed domain
reduce to the same reference-domain equation for matched initial and boundary data
and therefore generate identical pulled-back trajectories (Fig.~\ref{fig:confounding}b).
Although their geometries and physical-domain PDE coefficients differ before
pull-back, these quantities enter the trajectory through the same transformed
coefficient fields. We call this non-unique physical attribution
\emph{geometry--physics confounding}.

The information available for attribution and the chosen model class determine
whether this ambiguity remains. A misspecified reference-domain model that
compresses the transformed operator into a single apparent diffusion coefficient
yields identical fitted scalars ($\widehat\kappa_{\rm Ref}=0.010$) across distinct
geometry--coefficient pairs (Fig.~\ref{fig:confounding}b, right). That scalar
alone cannot determine how much of the observed rate of change originates from
geometric distortion and how much belongs to the material. Using the domain
map's PDE-specific action makes this decomposition available.
Figure~\ref{fig:confounding}c outlines the strategies detailed in the next subsection.

For operator learning, leaving that action implicit creates an
additional finite-data learning problem. The rate-prediction experiments below
focus on the geometry-dependent action that the learned predictor must
represent. We expanded Eq.~\eqref{eq:pullback-diffusion} into five state
derivatives and five geometry-induced coefficient fields derived from the
Jacobian. Their products form the five geometry-induced PDE terms used in the
final representation (Methods). The
representation sequence in Fig.~\ref{fig:confounding}d begins with
state and physical coordinates, followed by exact state derivatives, the
geometry-induced coefficient fields $q_g$ derived from the Jacobian, and the geometry-induced PDE terms
$q_g\odot\mathcal D[u]$.
The models using state derivatives and geometry-induced coefficient fields use the same neural
architecture and differ only in whether their inputs expose the
geometry-to-operator action. At the largest training sets, the latter reduced
test error by factors of 5.4 for pointwise rate prediction and 5.3 for
complete-field rate prediction at the held-out geometry
(Fig.~\ref{fig:confounding}d, left and centre).
Fitting independent linear weights to the five geometry-induced PDE terms
reduced both errors to numerical precision.

For equation discovery, an implicit geometric action
produces structural rather than only statistical failure. Before considering
the evolving-domain trajectories, the fixed-deformation experiment in
Fig.~\ref{fig:confounding}d (right) tests spatial candidate-library
misspecification. A discovery algorithm can select only operators present in its
candidate library. Data were generated with diffusion coefficient
$\kappa=0.01$ and fit with the standard reference Laplacian
$\Delta_\xi\widehat u$. As deformation increased, the apparent Laplacian
coefficient
drifted to $0.00807$, producing a $19.3\%$ bias
(Fig.~\ref{fig:confounding}d, top right). Because a
single scalar cannot reproduce the spatially varying anisotropic operator, the
relative equation residual simultaneously rose to $0.223$
(Fig.~\ref{fig:confounding}d, bottom right). The mapped diffusion operator recovered the
generating coefficient and reduced the structural residual to numerical
precision. The drift in the apparent Laplacian coefficient measures the
geometric effect absorbed by the misspecified term, while the remaining
residual measures the part that the library cannot represent. The
fixed-deformation and evolving-domain experiments test the same
candidate-library requirement. Time-dependent maps additionally introduce the
associated domain-motion operators.

\subsection*{De-confounding through geometry-induced representations}

The preceding analysis suggests a task-matched use of the known domain
correspondence. We use \emph{geometry-induced representations} for the
map-dependent quantities that arise when we write a physical-domain PDE on the
reference domain. The appropriate representation depends on the object
being learned. For operator learning, an equivalent weak-form pull-back provides
a systematic construction of the transformed problem specification. For
equation discovery, a strong-form pull-back maps the prespecified physical
library into structured reference-domain candidate operators. In both cases,
the representation makes the geometry-to-operator transformation explicit, so
the learning procedure need not reconstruct that transformation implicitly from
examples. Both constructions apply across related and evolving geometries, and a
time-dependent map additionally contributes the corresponding domain-motion
terms.

\paragraph*{Weak-form pull-back for operator learning.}
For $g\in\mathcal G$, suppose that the physical-domain problem is written in
the weak residual form
\begin{equation}
 \mathfrak R_{\domain_g(t)}(u,v;s)=0
 \quad\text{for all admissible }v.
 \label{eq:physical-weak-problem}
\end{equation}
This notation retains any time dependence. Subject to the regularity and
boundary-condition assumptions of the physical weak problem, pulling back the
trial and test fields and applying the change of variables to all volume and
boundary terms yields an equivalent problem on reference spaces,
\begin{equation}
 \widehat{\mathfrak R}
 \bigl(\widehat u,\widehat v;\widehat s^{\rm w}\bigr)=0
 \quad\text{for all }\widehat v\in\widehat{\mathcal V},
 \qquad
 \widehat s^{\rm w}=\mathcal W^{\rm w}(s;\map_g,\partial_t\map_g).
 \label{eq:reference-weak-problem}
\end{equation}
For operators that admit integration by parts under the imposed boundary
conditions, this weak-form construction also lowers the differentiation burden.
For example, a second-order divergence-form term can be evaluated after
pull-back using first derivatives of the trial and test fields together with
$j_g$, $\jac_g^{-1}$ and cofactor-like fields. It is then unnecessary to expand
the divergence pointwise and differentiate the map-induced coefficient
$j_g\jac_g^{-1}\jac_g^{-\mathsf T}$, an expansion that would generally require
second derivatives of the state and higher derivatives of the domain map. The
weak form can therefore avoid these higher-order evaluations while retaining an
equivalent variational problem. Here $\mathcal W^{\rm w}$ transforms the prescribed problem specification. It
acts jointly on the initial and boundary conditions, forcing terms, known PDE
coefficients and control inputs, as applicable, and includes all coefficient and
data fields generated by the changes of variables in the volume and boundary
terms. For a time-dependent map, these fields also include the map velocity and
associated domain-motion terms. We can therefore represent the varying-domain solution family on the
common reference domain through the solution operator
\begin{equation}
 \widehat{\mathcal F}:
 \widehat s^{\rm w}\longmapsto\widehat u,
 \qquad
 \mathcal F(s;\map_g)
 =\mathcal P_g^{-1}\widehat{\mathcal F}
 \bigl(\mathcal W^{\rm w}(s;\map_g,\partial_t\map_g)\bigr).
 \label{eq:reference-operator-learning}
\end{equation}
A physical-domain PDE coefficient that is constant may therefore become a
spatially varying coefficient field after pull-back. The operator-learning
formulation does not separate such a field into an invariant internal parameter
and a geometric factor. We treat it directly as part of the transformed problem
specification $\widehat s^{\rm w}$.

\paragraph*{Strong-form pull-back for equation discovery.}
For $g\in\mathcal G$, we specify the candidate library before
geometric transformation as
\begin{equation}
 \bm{\mathcal T}_{g,t}[u;s^{\rm known}]
 =\bigl(
 \mathcal T_{1,\domain_g(t)}[u;s^{\rm known}],\ldots,
 \mathcal T_{p,\domain_g(t)}[u;s^{\rm known}]
 \bigr).
 \label{eq:physical-candidate-library}
\end{equation}
Here $s^{\rm known}$ contains only the problem data supplied during
identification. The coefficients assigned to the candidates remain unknown.
Write $\widehat s^{\rm known}=\mathcal P_g^s s^{\rm known}$. Let
$\mathcal P_{g,t}^{\rm str}$ denote the strong-form change of variables induced
by $\map_g(\cdot,t)$. It acts on an entire differential expression. In
particular, the standard Eulerian time derivative obeys
\begin{equation}
 \left.\partial_t u\right|_x\circ\map_g
 =\left.\partial_t\widehat u\right|_\xi
 -\mathcal A_{g,t}[\widehat u],
 \qquad
 \mathcal A_{g,t}[\widehat u]
 :=\bigl(\jac_g^{-1}\partial_t\map_g\bigr)
 \cdot\nabla_\xi\widehat u,
 \label{eq:ale-pullback}
\end{equation}
where $\mathcal A_{g,t}$ is the arbitrary Lagrangian--Eulerian (ALE) transport
operator and the displayed sign follows the Eulerian convention
\cite{hughes1981ale,donea2004ale}. An evolving-domain construction includes
this term, while $\mathcal A_{g,t}=0$ for a time-independent map. More generally,
the domain-motion contribution is determined by the relative velocity between
the map and the adopted physical transport, and vanishes when the map follows
the material motion.
For a prescribed Eulerian description, the map fixes both the term and its sign,
so it can be collected in $\widehat{\mathcal Q}_{g,t}$ rather than fitted. A
material-time description may instead absorb the contribution. In the
identification experiments, we retain $\mathcal A_{g,t}$ as a
single geometry-induced candidate in a common transformed library. This choice
tests whether model selection retains explicit domain-motion transport when it
is active and rejects the same candidate when the adopted material description
makes its generating coefficient zero. We then replace each physical candidate
with its geometry-equivalent reference-domain operator and append the ALE
operator directly to the transformed library:
\begin{equation}
 \widehat{\mathcal T}_{k,g,t}
 :=\mathcal P_{g,t}^{\rm str}
 \bigl[\mathcal T_{k,\domain_g(t)}\bigr],
 \qquad
 \widehat{\bm\Theta}_{g,t}
 :=\bigl[
 \widehat{\mathcal T}_{1,g,t},\ldots,
 \widehat{\mathcal T}_{p,g,t},
 \mathcal A_{g,t}
 \bigr].
 \label{eq:geometry-complete-library}
\end{equation}
After applying the same transformation to the full governing balance, we
collect the known terms to obtain the reference-domain identification problem
\begin{equation}
 \widehat{\mathcal Q}_{g,t}
 [\widehat u;\widehat s^{\rm known}]
 =\widehat{\bm\Theta}_{g,t}
 [\widehat u;\widehat s^{\rm known}] \widetilde{\bm c}^{\star},
 \qquad g\in\mathcal G.
 \label{eq:reference-discovery-problem}
\end{equation}
Here $\widetilde{\bm c}^{\star}$ augments the full candidate coefficient vector
$\bm c^\star$ with the coefficient of the ALE candidate used in these
experiments. Both vectors retain zero entries for inactive candidates. Their
entries are weights of specified candidate operators, whose relation to physical
parameters depends on the chosen representation.
The ALE entry records whether domain-motion transport remains explicit under the
adopted time description, and its coefficient describes the adopted kinematic
representation. Each transformed physical candidate may combine several map-dependent
factors and reference derivatives, but it remains a single structured operator
with one fitted coefficient.

%% file: sections/results/03_01_neural_operator.tex
\newsavebox{\operatorfigurecaptionbox}
\newlength{\operatorfigureimageheight}
\newcommand{\operatorfigurepage}[3]{%
  \begin{figure}[p]
    \centering
    \sbox{\operatorfigurecaptionbox}{%
      \begin{minipage}{\linewidth}
        \caption{#2}\label{#3}
      \end{minipage}%
    }%
    \setlength{\operatorfigureimageheight}{\dimexpr
      \textheight-\ht\operatorfigurecaptionbox-\dp\operatorfigurecaptionbox-2pt\relax}%
    \makebox[\linewidth][c]{%
      \vbox{%
        \offinterlineskip
        \hbox to \linewidth{\hfil
          \includegraphics[width=\linewidth,height=\operatorfigureimageheight,keepaspectratio]{#1}%
          \hfil}%
        \copy\operatorfigurecaptionbox
      }%
    }%
  \end{figure}%
}

\clearpage
\operatorfigurepage{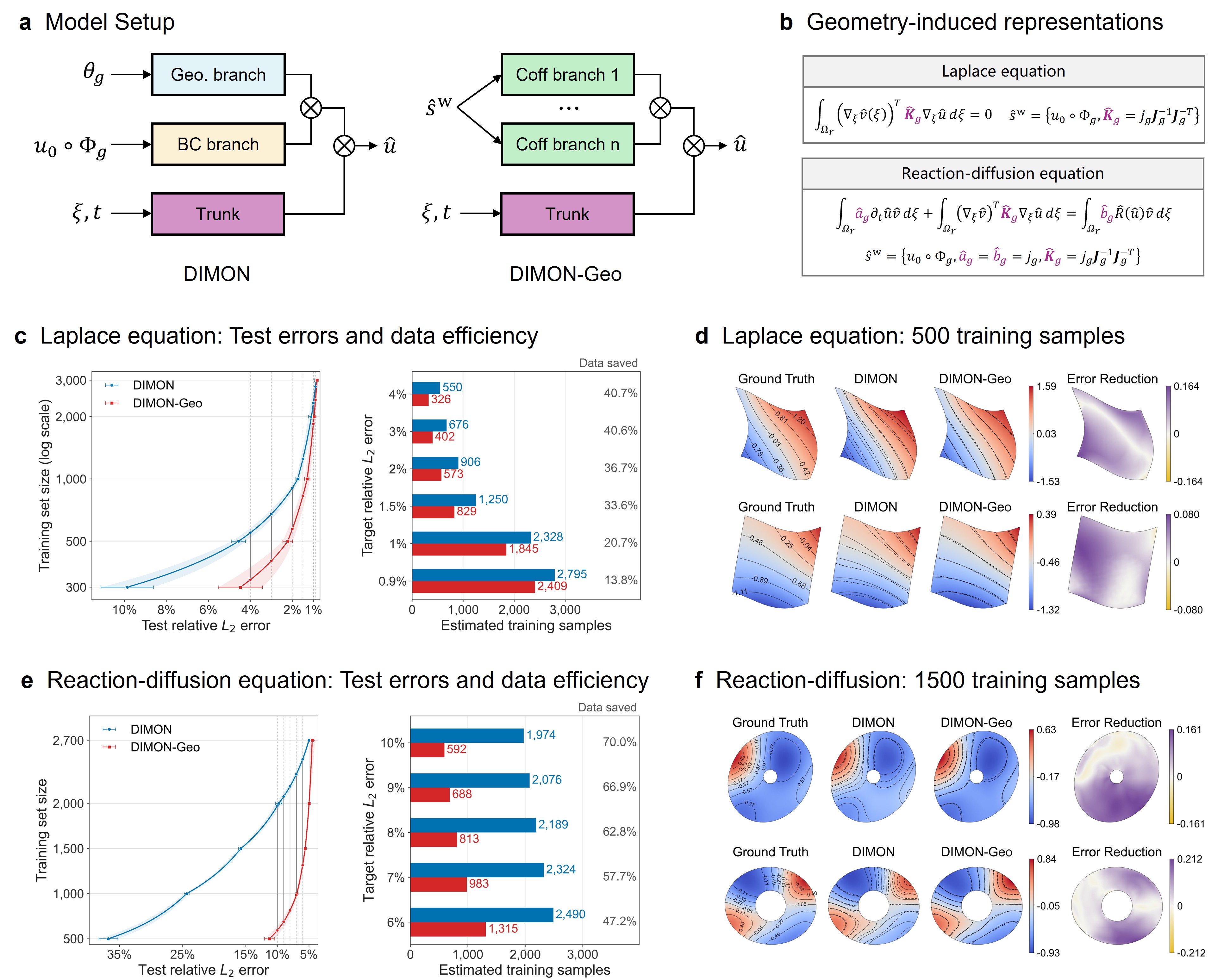}{%
  \textbf{Prediction accuracy and data efficiency of DIMON and DIMON-Geo.}
  \textbf{a}, Model configurations.
  \textbf{b}, Geometry-induced weak-form representations for Laplace and reaction--diffusion.
  \textbf{c,e}, Test relative $L_2$ errors versus training-set size (left) and estimated sample requirements at matched target errors (right) for Laplace and reaction--diffusion, respectively.
  \textbf{d}, Laplace predictions for two test samples using 500 training samples.
  \textbf{f}, Reaction--diffusion predictions for two test samples at $t=0.16$ (top) and $t=0.32$ (bottom), using 1500 training samples.
  Examples in \textbf{d,f} use seed 0.
  In \textbf{c,e}, markers and error bars show the mean and $\pm1$ sample s.d.\ of test-set mean errors across five seeds. Curves and shaded bands show the interpolated mean and mean $\pm1$ sample s.d., respectively. Sample requirements are estimated by interpolation (Methods), with percentages indicating savings relative to DIMON.
  In \textbf{d,f}, pointwise error reduction is DIMON absolute error minus DIMON-Geo absolute error, so positive values favour DIMON-Geo. Solid and dashed contours denote ground truth and predictions, respectively.}
  {fig:operator-dimon}
\clearpage

\section*{Results}
\subsection*{Geometry-induced representations improve neural-operator predictions across varying geometries}

We tested whether explicitly representing geometry-induced operator variation
improves prediction within existing neural-operator formulations.
The five benchmarks comprised the Laplace and nonlinear reaction--diffusion
examples from DIMON \cite{yin2024dimon}, and the public pipe-flow, airfoil and
elasticity benchmarks introduced with Geo-FNO and subsequently evaluated with
Transolver \cite{li2023geofno,wu2024transolver}. They span steady and
time-dependent equations, fluid and solid physics, and structured and
unstructured discretizations. Within each benchmark, the model architectures
and training settings were matched as closely as possible between the two
variants. Differences concerned the input representations and the associated
adaptations to input processing and network architecture, as detailed in
Supplementary Sections S2--S6.
We varied the number of labelled geometry--solution samples and repeated
each configuration with five paired random seeds. Reported errors are the
mean test relative $L_2$ error $\pm$ one sample standard deviation across
these runs. Data-efficiency estimates refer to labelled solutions.

\paragraph*{Reference-domain mapping models.}
DIMON learns geometry-dependent solution maps on a common reference domain.
DIMON-Geo retains its branch--trunk formulation and supplies geometric
coefficient fields through dedicated branches alongside the boundary or
initial data (Fig.~\ref{fig:operator-dimon}a). The weak-form motivation is
summarised in Fig.~\ref{fig:operator-dimon}b and derived in Methods.
The reaction--diffusion variant also changes the initial-condition encoding
and branch capacity (Supplementary Section S3). DIMON-Geo achieved lower mean test relative
$L_2$ errors at every tested training size in both problems
(Fig.~\ref{fig:operator-dimon}c,e).

For Laplace, at $n=300$, DIMON-Geo reduced the error from
$0.0987\pm0.0124$ to $0.0448\pm0.0105$ (54.6\%). At $n=3000$, the reduction
was smaller, from $0.00864\pm0.00090$ to $0.00827\pm0.00037$ (4.3\%).
At a target relative error of 2\%, interpolation of the mean learning curves
estimated 906 training samples for DIMON and 573 for DIMON-Geo, a 36.7\%
reduction. Estimated sample savings ranged from 13.8\% to 40.7\% across the
plotted target errors (Fig.~\ref{fig:operator-dimon}c). Figure~\ref{fig:operator-dimon}d
compares predictions and pointwise absolute-error reductions for two test
samples.

For reaction--diffusion, the reduction increased from 10.4\% at $n=2700$
($0.0501\pm0.0016$ to $0.0449\pm0.0045$) to 69.4\% at $n=500$
($0.3676\pm0.0149$ to $0.1127\pm0.0078$). At a target relative error of
10\%, the estimated training requirement decreased from 1974 samples for DIMON
to 592 for DIMON-Geo, a 70.0\% reduction. Estimated sample savings ranged from
47.2\% to 70.0\% across the plotted target errors
(Fig.~\ref{fig:operator-dimon}e). Figure~\ref{fig:operator-dimon}f shows two
illustrative test-sample snapshots. In both problems, the geometry-conditioned
variant showed its largest relative error reduction at the smallest tested
training size.

\operatorfigurepage{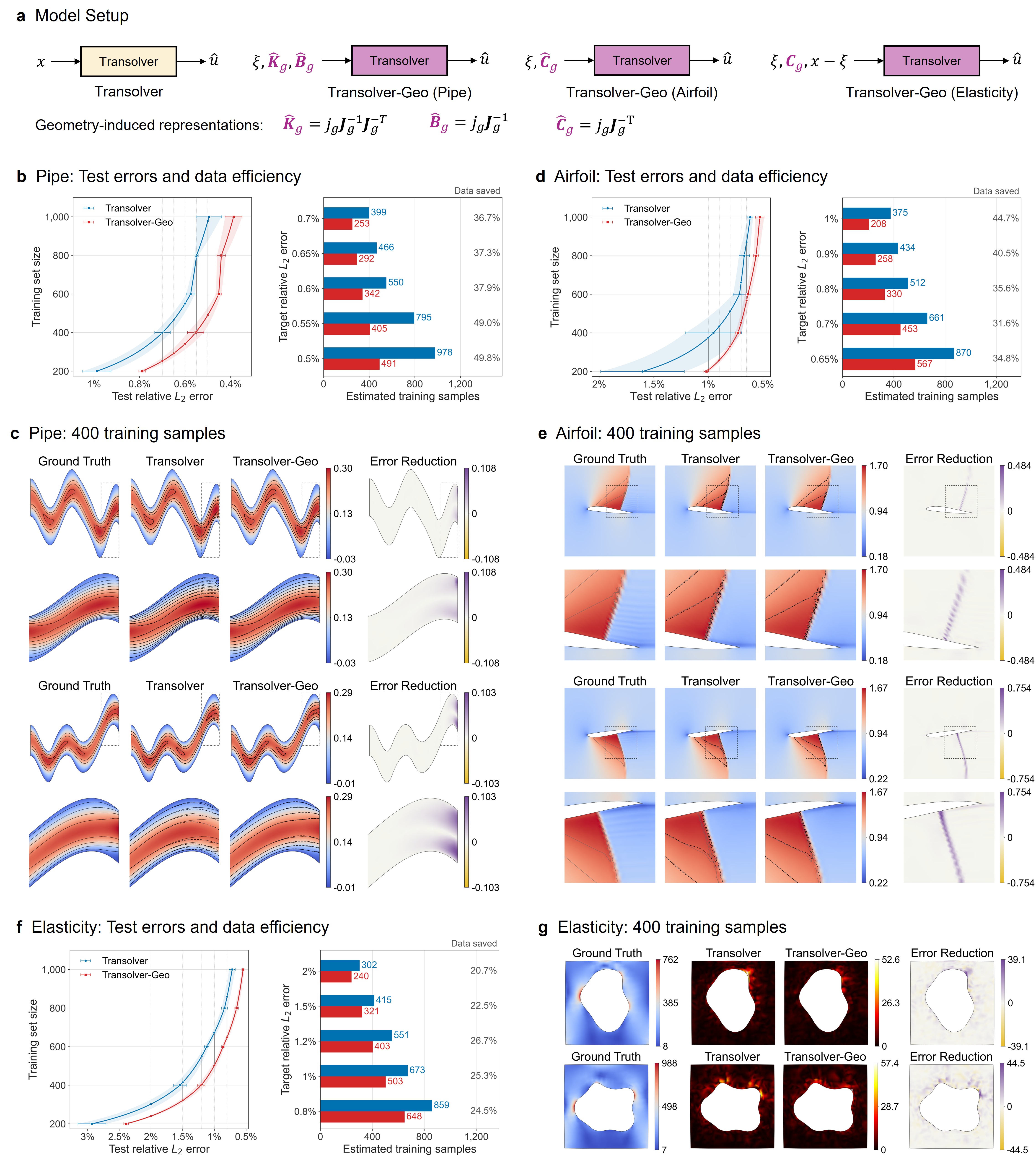}{%
  \textbf{Prediction accuracy and data efficiency of Transolver and Transolver-Geo.}
  \textbf{a}, Model configurations and geometric inputs.
  \textbf{b,d,f}, Test relative $L_2$ errors versus training-set size (left) and estimated sample requirements at matched target errors (right) for pipe flow, airfoil flow and elasticity, respectively.
  \textbf{c,e}, Horizontal-velocity and Mach-number predictions for pipe and airfoil flow, with full fields and outlet- or shock-region enlargements, respectively.
  \textbf{g}, Elastic-stress comparisons, with absolute errors in the two model columns.
  In \textbf{c,e,g}, two test samples per case are shown using 400 training samples and seed 0.
  In \textbf{b,d,f}, markers and error bars show the mean and $\pm1$ sample s.d.\ of test-set mean errors across five seeds. Curves and shaded bands show the interpolated mean and mean $\pm1$ sample s.d., respectively. Sample requirements are estimated by interpolation (Methods), with percentages indicating savings relative to Transolver.
  In \textbf{c,e,g}, pointwise error reduction is Transolver absolute error minus Transolver-Geo absolute error, so positive values favour Transolver-Geo.}
  {fig:operator-transolver}

\paragraph*{Geometry-aware attention models.}
We next evaluated the same representation principle in Transolver
(Fig.~\ref{fig:operator-transolver}a). Within each benchmark, the
baseline and geometry-conditioned variants shared the attention backbone and
case-specific optimisation schedule. Transolver-Geo replaces physical-coordinate
inputs with fixed reference coordinates and case-specific geometric channels.
Field definitions are given in Methods and implementation details in Supplementary
Sections S4--S6.
The reference meshes use all 1,000 training-pool geometries, including for
smaller labelled subsets. Under these protocols, 
Transolver-Geo achieved lower mean test relative
$L_2$ errors at every tested training size in all three benchmarks
(Fig.~\ref{fig:operator-transolver}b,d,f).

For pipe flow, the relative reduction remained between 19.8\% and 21.8\%
across the learning curve. At $n=200$, the error decreased from
$0.00988\pm0.00062$ to $0.00789\pm0.00015$. With 600 samples, the augmented
model attained a lower error than the baseline with 1000 samples (0.00452
versus 0.00495). Across the target errors plotted in
Fig.~\ref{fig:operator-transolver}b, the estimated reduction in required
training samples ranged from 36.7\% to 49.8\%. Figure~\ref{fig:operator-transolver}c
compares the full horizontal-velocity fields and outlet-region enlargements
for two test samples.

For airfoil flow, Transolver-Geo reduced the mean error by 36.5\% at $n=200$
($0.0160\pm0.0038$ to $0.01018\pm0.00024$), compared with 14.1\% at
$n=1000$. The augmented model with 800 samples was also more accurate than the
baseline with 1000 samples (0.00563 versus 0.00616).
Across the plotted target errors, the estimated sample savings ranged from
31.6\% to 44.7\% (Fig.~\ref{fig:operator-transolver}d).
Figure~\ref{fig:operator-transolver}e compares the full Mach-number fields and
shock-region enlargements for two test samples.

For elasticity, the target is the stress field in a unit cell with a varying
internal void. The response depends on the constitutive law and the domain
and boundary configuration, whereas $C_g$ describes the local stress-flux
transformation rather than a complete elasticity operator. We therefore
supplement $C_g$ with the domain-map displacement $d_g=x-\xi$, which makes
the global shape and boundary positions explicit relative to the reference
mesh. This field describes the input geometry, not the mechanical displacement
induced by loading. The variant using these joint $C_g+d_g$ inputs reduced the mean error by
18.4\% at $n=200$ and 24.1\% at $n=1000$. With 800 samples, it outperformed
the baseline trained with 1000 samples (0.00649 versus 0.00720). Across the
plotted target errors, the estimated sample savings ranged from 20.7\% to
26.7\% (Fig.~\ref{fig:operator-transolver}f). Figure~\ref{fig:operator-transolver}g
shows the ground-truth stress, the absolute errors of the two models and their
pointwise error reduction for two test samples. The improvement is consistent
with the complementary geometric information provided by the domain-map
displacement and the cofactor field.

Across the five benchmarks, the geometry-conditioned variants achieved lower
mean prediction errors at all tested training sizes under the case-specific
protocols. Together, the learning
curves support the predictive utility and data efficiency of these
representations within the tested model families, equations and discretizations.

%% file: sections/results/03_02_mechanism_identification.tex
\subsection*{Geometry-complete libraries enable equation discovery on evolving domains}

Equation discovery was evaluated for deformation-coupled transport and
growth-driven free-boundary transport
\cite{friedman2015freeboundary,long2026freeboundary}. In both systems, the
geometry trajectory and its evolution closure were known, and the identified
state equation was coupled to that closure for held-out prediction. We compare
a reference-coordinate library (Ref), containing the standard operators defined
on the fixed reference domain, with a geometry-complete library (Geo), which
retains these terms and adds the mapped diffusion and ALE transport candidates
induced by the observed geometry. The two candidate libraries are compared in
Extended Data Fig.~1. Extended Data Fig.~2 shows the training and held-out
geometries.

\begin{figure}[t!]
  \centering
  \maybeincludegraphics[width=0.99\linewidth]{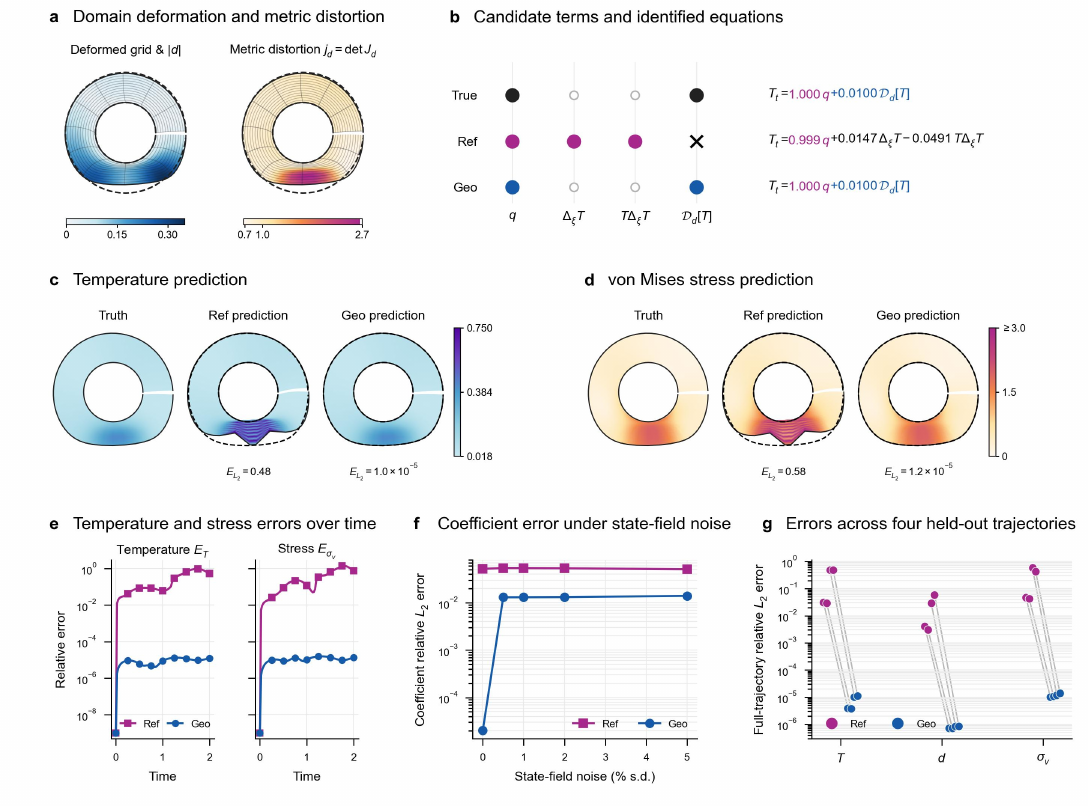}
  \caption{\textbf{Thermoelastic equation discovery and prediction on a held-out realization.}
  \textbf{a}, Domain deformation $|d|$ and metric distortion $j_d=\det J_d$ for a representative held-out realization.
  \textbf{b}, Generating and selected terms and the corresponding equations identified using the reference-coordinate (Ref) and geometry-complete (Geo) candidate libraries. Filled and open circles denote included and excluded terms, respectively, and the cross marks the generating term missed by Ref. In the equations, purple denotes the source term $q$ and its fitted coefficient, blue denotes the mapped-diffusion term and its fitted coefficient, and black denotes the additional terms selected by Ref.
  \textbf{c}, Temperature prediction on a representative held-out trajectory at $t=2.0$, comparing ground truth with predictions from the reference-coordinate and geometry-complete identified equations. Dashed outlines mark the true boundary.
  \textbf{d}, Coupled von Mises stress prediction computed from the predicted temperature and displacement fields.
  \textbf{e}, Relative errors in temperature and von Mises stress over time.
  \textbf{f}, Coefficient relative $L_2$ error under 0--5\% additive state-field noise.
  \textbf{g}, Full-trajectory relative $L_2$ error across four held-out trajectories for temperature $T$, displacement $\bm d$, and von Mises stress $\sigma_v$. The geometry closure is known.}
  \label{fig:thermal-discovery}
\end{figure}

\paragraph*{Heat transport on a deforming solid.}
The thermoelastic case tests whether a single intrinsic conduction law can be
identified while thermal expansion changes the physical domain. Let $T$ denote
temperature, $q$ the applied heat source and $\bm d$ the displacement defining
$x=\xi+\bm d(\xi,t)$. On reference coordinates, the energy balance is
\begin{equation}
 T_t=0.01\,\mathcal D_d[T]+q,
 \qquad
 \mathcal D_d[T]=j_d^{-1}\operatorname{Div}_\xi
 \left(j_d\jac_d^{-1}\jac_d^{-\mathsf T}\nabla_\xi T\right).
 \label{eq:thermal-generating-v2}
\end{equation}
Figure~\ref{fig:thermal-discovery}a shows the displacement magnitude and metric
distortion for a representative held-out realization. The non-uniform
$j_d$ field illustrates the spatially varying geometric action contained in
$\mathcal D_d[T]$.

The reference-coordinate candidate library was structurally misspecified because
it did not contain the mapped diffusion operator $\mathcal D_d[T]$. The selected
equation therefore combined reference diffusion with a spurious
state-modulated term $T\Delta_\xi T$, giving a support $F_1$ score of $0.4$ and a
held-out PDE residual of $0.171$ (Fig.~\ref{fig:thermal-discovery}b). In
contrast, the geometry-complete candidate library recovered the generating
terms $q$ and $\mathcal D_d[T]$, rejected both inactive ALE candidates, and
estimated the generating candidate coefficients with a relative error of
$2.03\times10^{-5}$. Their rejection is consistent with the material-time
description adopted for this system, which assigns zero generating
coefficients to domain-motion transport. The held-out PDE residual was
$3.02\times10^{-5}$, more than three orders of magnitude lower than that of
the reference-coordinate candidate library.

Recovering the generating equation also reduced closed-loop prediction errors.
On a representative held-out realization, spatial predictions at $t=2.0$ show
that the reference-coordinate model incurs severe localized errors along the
deformed inner boundary ($E_{L_2}=0.48$ for temperature, Fig.~\ref{fig:thermal-discovery}c, and $0.58$ for von Mises
stress, Fig.~\ref{fig:thermal-discovery}d), whereas the geometry-complete equation matches the true boundary
profile with relative errors below $1.2\times10^{-5}$.
In forward integration, reference errors rise rapidly within the initial
steps to order $10^0$, whereas geometry-complete errors remain bounded
below $10^{-5}$ over time (Fig.~\ref{fig:thermal-discovery}e).
Across four held-out trajectories with shifted heat sources, the geometry-complete
equation consistently reduces full-trajectory relative $L_2$ errors in
temperature $T$, displacement $\bm d$, and von Mises stress $\sigma_v$ by more
than four orders of magnitude relative to the reference-coordinate baseline
(Fig.~\ref{fig:thermal-discovery}g and Extended Data Fig.~3).

\FloatBarrier

\begin{figure}[t!]
  \centering
  \maybeincludegraphics[width=0.99\linewidth]{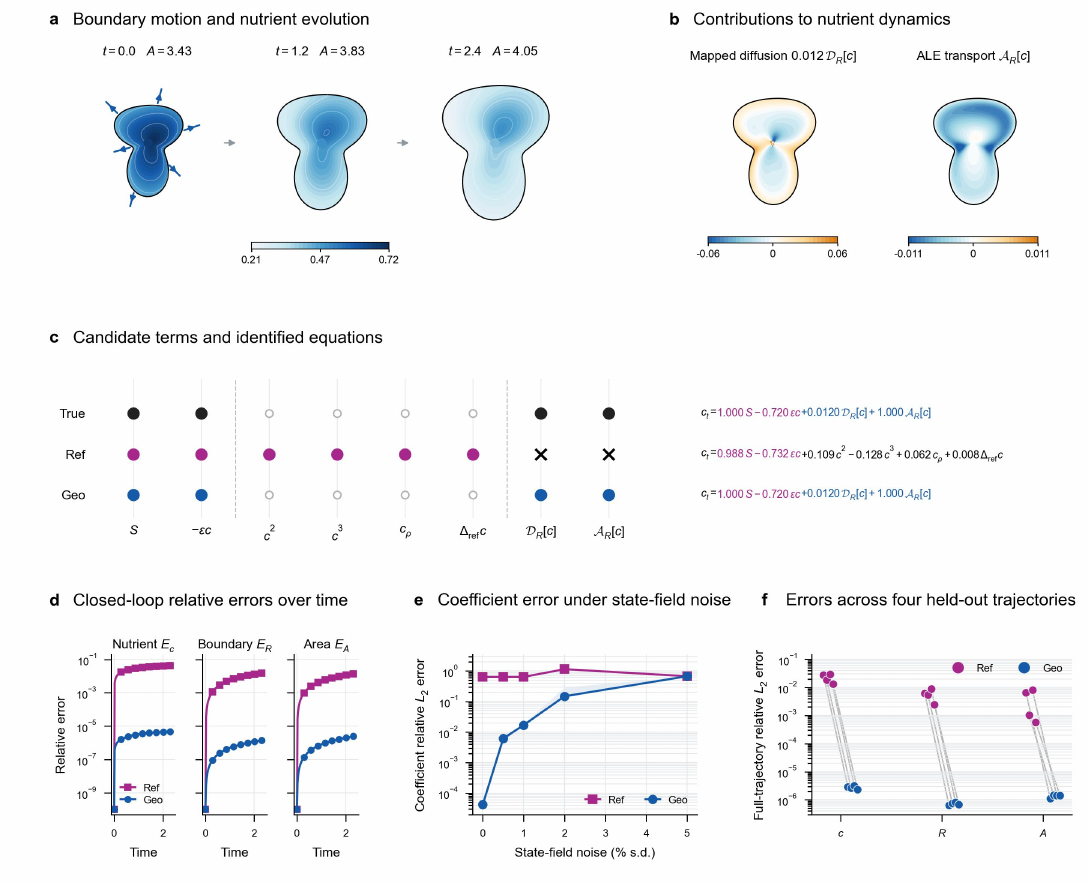}
  \caption{\textbf{Tumour-growth equation discovery and prediction on a held-out realization.}
  \textbf{a}, Free-boundary motion and nutrient evolution on a shared colour scale. The domains are uniformly rescaled within each snapshot to make growth visible. Arrows indicate the initial direction of boundary motion and labels report physical domain area $A$.
  \textbf{b}, Generating mapped-diffusion ($0.012\mathcal D_R[c]$) and ALE-transport ($\mathcal A_R[c]$) PDE contributions at $t=1.2$.
  \textbf{c}, Generating and selected terms and the corresponding identified equations. Filled and open circles denote included and excluded terms, respectively, and crosses mark the generating terms missed by Ref. In the equations, purple denotes the source and consumption terms and their fitted coefficients, blue denotes the mapped-diffusion and ALE-transport terms and their fitted coefficients, and black denotes the additional terms selected by Ref.
  \textbf{d}, Closed-loop relative errors over time for nutrient $E_c$, boundary radius $E_R$, and domain area $E_A$.
  \textbf{e}, Coefficient relative $L_2$ error under 0--5\% state-field noise.
  \textbf{f}, Full-trajectory relative $L_2$ error across four held-out trajectories for nutrient $c$, boundary $R$, and area $A$. The moving-boundary closure is known.}
  \label{fig:tumour-discovery}
\end{figure}

\paragraph*{Nutrient transport on a growing domain.}
The tumour-growth case tests the joint identification of mapped diffusion and
transport induced by boundary motion. Let $c$ denote nutrient concentration,
$R(\theta,t)$ the boundary radius, $S$ nutrient supply and $\epsilon$ the
metabolic uptake field. On reference polar coordinates, the transport law is
\begin{equation}
 c_t=0.012\,\mathcal D_R[c]+\mathcal A_R[c]+S-0.72\,\epsilon c,
 \qquad
 \mathcal A_R[c]=\rho\frac{R_t}{R}c_\rho .
 \label{eq:tumour-generating-v2}
\end{equation}
The representative trajectory in Fig.~\ref{fig:tumour-discovery}a shows the
increase in domain area from $3.43$ to $4.05$ together with the redistribution
of nutrient concentration. At $t=1.2$, mapped diffusion and ALE transport have
distinct spatial contributions to the nutrient dynamics
(Fig.~\ref{fig:tumour-discovery}b).

In this setting, the reference-coordinate candidate library omitted both mapped
diffusion $\mathcal D_R[c]$ and ALE transport $\mathcal A_R[c]$. The selected
equation therefore failed to recover the mapped-diffusion and ALE-transport
terms and included four non-generating terms, giving an $F_1$ score of $0.4$ and an equation residual of
$0.180$ (Fig.~\ref{fig:tumour-discovery}c). The geometry-complete candidate
library recovered the four generating terms with a candidate-coefficient relative error
of $4.25\times10^{-5}$ and an equation residual of $6.93\times10^{-4}$, more
than two orders of magnitude lower than the reference-coordinate residual.

During closed-loop evolution, the reference-coordinate model accumulates errors
rapidly, reaching relative errors of order $10^{-1}$ in nutrient concentration $E_c$,
boundary radius $E_R$ and domain area $E_A$ (Fig.~\ref{fig:tumour-discovery}d).
By contrast, the geometry-complete equation maintains accurate tracking throughout
the trajectory, keeping all three relative errors bounded between $10^{-6}$ and
$10^{-5}$. Across four held-out growth trajectories with non-convex, multi-lobed
boundaries, the geometry-complete equation consistently reduces full-trajectory
relative $L_2$ errors in nutrient concentration, boundary radius and total
domain area by three to four orders of magnitude relative to the
reference-coordinate baseline
(Fig.~\ref{fig:tumour-discovery}f and Extended Data Fig.~4).

\paragraph*{Recovery under noise and limited trajectory diversity.}
The sensitivity of numerical differentiation to measurement
noise was next evaluated after supplying the required candidate operators. With additive state-field noise
and noise-free geometry, the thermoelastic system retained exact support at all
tested levels up to $5\%$. At $5\%$, the median coefficient error was $0.0141$
and the held-out residual was $0.0143$
(Fig.~\ref{fig:thermal-discovery}f). The tumour-growth system retained exact
support up to $1\%$ noise, whereas the median $F_1$ score decreased to $0.727$
at $2\%$ and $0.444$ at $5\%$ (Fig.~\ref{fig:tumour-discovery}e). This pattern
is consistent with the numerical sensitivity of the coupled
advection--diffusion operators (Extended Data Fig.~5). In the clean-data
trajectory-subset analysis, one trajectory recovered the generating support in both
systems, while pooling trajectories reduced coefficient variability across the
enumerated subsets (Extended Data Fig.~6). These results separate
candidate-library misspecification from the effects of state-field noise and
training-trajectory selection in the tested systems.

%% file: sections/05_discussion.tex
Geometry--physics confounding links two apparently different limitations of PDE
learning across varying domains. In operator learning, an implicit geometric
action must be approximated from finite examples and therefore increases the
representation burden. In equation discovery, the same omission changes the
candidate space itself: an absent transformed operator cannot be recovered by
sparse regression. Domain motion extends this distinction by introducing a
transport contribution whose presence depends on the adopted physical
description.

The operator-learning results show that geometry-induced coefficient fields can
reduce this representation burden within existing solution-operator
formulations. The improvement across DIMON and Transolver indicates that the useful
information is not tied to a particular architecture or discretisation. Its
effect nevertheless depends on how completely the supplied fields describe the
transformed PDE. For example, elasticity requires both the local cofactor field
and the displacement field to represent local deformation and global shape.

For equation discovery, the fixed-deformation and evolving-domain results show
the corresponding structural effect. Reference-coordinate libraries absorbed
unrepresented geometric action into biased coefficients and additional terms,
whereas the geometry-complete libraries recovered candidate coefficients shared across
training trajectories. The thermoelastic system rejected inactive ALE
candidates under its material-time description, while the tumour system
retained active domain-motion transport. The noise results further show that
candidate completeness and derivative quality are separate requirements:
supplying the correct operators removes library misspecification but does not
eliminate sensitivity to noisy state derivatives.

The present cases primarily involve relatively ideal geometric maps.
Accurate, well-conditioned mappings are expected to provide the most favourable
setting for geometry-induced representations by limiting errors in the
transformed coefficients. Nevertheless, we also observed benefits for PDE
learning in datasets containing some lower-quality mappings, suggesting that
these representations can remain useful without uniformly ideal geometric
correspondences. This motivates further relaxing the requirement for exact
geometry-induced terms and examining how approximate representations affect
learning performance. Future work will assess sensitivity to mapping and
representation errors and seek to establish the
conditions under which geometry-induced representations improve PDE learning
and the extent of their theoretical advantages through formal analysis.

%% file: sections/methods/04_01_geometry_and_diagnostic.tex
\subsection*{Domain maps and geometry-induced representation}

We use the domain correspondence $x=\map_g(\xi,t)$, its Jacobian
$\jac_g$ and determinant $j_g>0$ defined in Eq.~\eqref{eq:reference-map}.
The pull-back of a state or test field is denoted by a hat, as in
$\widehat u=u\circ\map_g$. The geometric action is represented by coefficient
fields for operator learning and by assembled differential terms for equation
discovery. The constructions below specify these two representations.

\paragraph*{Geometry-induced fields for operator learning.}
The five solution-operator benchmarks use time-independent domain maps.
For each case, we give the equivalent weak
problem on $\refdomain$ and identify the geometric fields that enter it.
Trial fields satisfy any strongly imposed boundary conditions, and the
corresponding test fields vanish on those boundaries. Each weak identity
holds for all admissible test fields with sufficient regularity for the
displayed integrals to be well defined. Scalar gradients are column vectors.
For vector fields, gradient rows correspond to field components. Thus
$(\nabla_xu)\circ\map_g=\jac_g^{-\mathsf T}\nabla_\xi\widehat u$,
$(\nabla_x\bm u)\circ\map_g=(\nabla_\xi\widehat{\bm u})\jac_g^{-1}$
and $dx=j_g\,d\xi$.

\textit{Laplace.}
For the Laplace equation $-\Delta_xu=0$ with prescribed boundary values
$u=u_D$, the equivalent weak problem is to find $\widehat u$
with boundary trace $u_D\circ\map_g$ such that
\begin{equation}
 \int_{\refdomain}(\nabla_\xi\widehat v)^{\mathsf T}
       \keff_g\nabla_\xi\widehat u\,d\xi=0,
 \label{eq:weak-laplace}
\end{equation}
for all scalar test fields $\widehat v$ vanishing on $\partial\refdomain$.
Here $\keff_g=j_g\jac_g^{-1}\jac_g^{-\mathsf T}$ is the geometric diffusion
factor.
Geometry enters the interior balance through $\keff_g$ and the prescribed
data through the boundary map. DIMON-Geo therefore supplies the three
independent entries of this symmetric two-dimensional tensor and the
pulled-back boundary data through separate branches.

\textit{Reaction--diffusion.}
For the the Reaction--diffusion equation $u_t=\Delta_xu+r(u)$, where $r(u)=u(1-u^2)$
\cite{yin2024dimon}, zero normal diffusive flux gives the equivalent weak
balance
\begin{equation}
 \int_{\refdomain}j_g\partial_t\widehat u\,\widehat v\,d\xi
 +\int_{\refdomain}(\nabla_\xi\widehat v)^{\mathsf T}
       \keff_g\nabla_\xi\widehat u\,d\xi
 =\int_{\refdomain}j_gr(\widehat u)\widehat v\,d\xi,
 \label{eq:weak-reaction-diffusion}
\end{equation}
for every admissible scalar test field $\widehat v$.
The initial condition is $\widehat u(\xi,0)=u_0\circ\map_g(\xi)$.
Accordingly, DIMON-Geo supplies $j_g$ and the three independent entries of
$\keff_g$, together with the pulled-back initial condition. The reaction
function retains its form, but its weak integral carries the volume weight
$j_g$, as does the temporal term.

\textit{Pipe flow.}
Let $\bm v$ and $p$ denote velocity and pressure per unit density. The steady
incompressible equations are
$(\bm v\cdot\nabla_x)\bm v-\nu\Delta_x\bm v+\nabla_xp=0$ and
$\nabla_x\cdot\bm v=0$, with constant kinematic viscosity $\nu$.
For prescribed inlet velocity, no-slip walls and the homogeneous natural
outlet condition $(\nu\nabla_x\bm v-p\bm I)\bm n_x=0$, their equivalent
weak form is
\begin{equation}
 \begin{aligned}
 &\int_{\refdomain}
   \bigl[(\nabla_\xi\widehat{\bm v})B_g\widehat{\bm v}\bigr]
   \cdot\widehat{\bm V}\,d\xi
 +\nu\int_{\refdomain}
   \bigl[(\nabla_\xi\widehat{\bm v})\keff_g\bigr]:
   \nabla_\xi\widehat{\bm V}\,d\xi\\
 &\hspace{3em}
 -\int_{\refdomain}\widehat p\,C_g:
   \nabla_\xi\widehat{\bm V}\,d\xi=0,\\
 &\int_{\refdomain}\widehat\psi\,C_g:
   \nabla_\xi\widehat{\bm v}\,d\xi=0,
 \end{aligned}
 \label{eq:weak-pipe}
\end{equation}
for all vector tests $\widehat{\bm V}$ vanishing on the inlet and walls and
scalar pressure tests $\widehat\psi$. The flux factors are
$B_g=j_g\jac_g^{-1}$ and
$C_g=j_g\jac_g^{-\mathsf T}=B_g^{\mathsf T}=\cofj_g$.
Thus convection uses $B_g$, viscous diffusion uses $\keff_g$, and pressure
and incompressibility use $C_g=B_g^{\mathsf T}$. The four entries of $B_g$
and the three independent entries of $\keff_g$ give the seven geometric
feature channels, without a separate set of channels for $C_g$.
The prediction target is the horizontal Cartesian velocity component
$v_x=\bm v\cdot\bm e_x$. Pressure enters the governing momentum balance
used to derive these fields, but is not the learned output in this benchmark.

\textit{Airfoil flow.}
For the steady compressible Euler system
$\nabla_x\cdot\bm F(\bm U)=0$, let $\bm U$ be the conservative state and
let the rows of $\bm F(\bm U)$ contain its mass, momentum and energy flux
vectors. Integration by parts and pull-back give the equivalent weak balance
\begin{equation}
 \int_{\refdomain}
   \bigl[\bm F(\widehat{\bm U})C_g\bigr]:
   \nabla_\xi\widehat{\bm V}\,d\xi
 =\int_{\partial\refdomain}
   \bigl[\bm F(\widehat{\bm U})C_g\bm n_\xi\bigr]
   \cdot\widehat{\bm V}\,dS_\xi
 \label{eq:weak-euler}
\end{equation}
for every admissible vector test field $\widehat{\bm V}$, with the wall
and far-field conditions imposed in the boundary flux. Here $\bm n_\xi$
is the outward unit normal, and the transformed surface vector satisfies
$\bm n_x\,dS_x=C_g\bm n_\xi\,dS_\xi$. Hence the same cofactor field
acts on the volume and boundary fluxes. Its four entries are the geometric
feature channels for this steady problem. No temporal volume-weight term is
present.

\textit{Elasticity.}
For static hyperelasticity, let $\bm u$ be the mechanical displacement on
the physical undeformed domain $\domain_g$, and let $\bm\Sigma$ denote
the first Piola--Kirchhoff stress. With no body force, equilibrium is
$\nabla_x\cdot\bm\Sigma=0$. If $\bm t_N=\bm\Sigma\bm n_x$ is the
prescribed traction on $\Gamma_N$, the equivalent weak problem is
\begin{equation}
 \int_{\refdomain}(\widehat{\bm\Sigma}C_g):
       \nabla_\xi\widehat{\bm V}\,d\xi
 =\int_{\widehat\Gamma_N}
       \|C_g\bm n_\xi\|\,\widehat{\bm t}_N
       \cdot\widehat{\bm V}\,dS_\xi,
 \label{eq:weak-elasticity}
\end{equation}
for all vector tests $\widehat{\bm V}$ vanishing on the prescribed-displacement boundary,
where $\widehat\Gamma_N=\map_g^{-1}(\Gamma_N)$. The surface factor
$\|C_g\bm n_\xi\|$ converts traction per unit physical boundary measure.
For strain-energy density $W$, the pulled-back constitutive relation is
\begin{equation}
 \widehat{\bm F}_m
 =\bm I+(\nabla_\xi\widehat{\bm u})\jac_g^{-1},\qquad
 \widehat{\bm\Sigma}
 =\frac{\partial W}{\partial\bm F_m}(\widehat{\bm F}_m),
 \label{eq:weak-elasticity-constitutive}
\end{equation}
where $\bm F_m=\bm I+\nabla_x\bm u$ is the material deformation gradient,
distinct from the domain-map Jacobian $\jac_g$. The weak form identifies
$C_g$ as the stress-flux factor, while the stress itself also depends on
the unknown displacement through the transformed constitutive law. Thus
$C_g$ is not, by itself, the assembled elasticity operator. The implemented
input pairs its four entries with the
two-component domain-map displacement $d_g=x-X$, where $X$ denotes the
corresponding reference point. This additional field retains the global shape
and boundary position. It is supplied geometric information, not the
mechanical response being predicted.

\paragraph*{Geometry-induced operator terms for equation discovery.}
For equation discovery, geometry-dependent coefficient fields are combined
with state derivatives to form complete reference-domain candidate terms.
We use pointwise strong-form pull-backs rather than the weak integrals above.
The weak-integration volume weight is therefore not appended to each term,
although factors such as $j_g$ remain within the transformed operators.

For a physical vector flux $\bm q$, the Piola identity
reads $(\nabla_x\cdot\bm q)\circ\map_g
=j_g^{-1}\nabla_\xi\cdot(B_g\widehat{\bm q})$,
where $\widehat{\bm q}=\bm q\circ\map_g$. Together with the gradient
transformation above, this identity yields the mapped diffusion term.
Table~\ref{tab:geometry-induced-catalogue} summarises common transformation
rules, rather than a complete regression library. Here
$\widehat{\bm b}=\bm b\circ\map_g$ is a prescribed advective velocity
pulled back to the reference domain, and $s$ is a physical-domain source.

\begin{table}[H]
\centering
\caption{Common PDE terms and their geometry-induced forms.}
\label{tab:geometry-induced-catalogue}
\footnotesize
\begin{tabular}{@{}>{\raggedright\arraybackslash}p{2.35cm}
                    >{\raggedright\arraybackslash}p{2.7cm}
                    >{\raggedright\arraybackslash}p{6.75cm}
                    >{\centering\arraybackslash}p{1.6cm}@{}}
\toprule
Term & Physical-domain form & Geometry-induced form & Short notation \\
\midrule
Gradient & $\nabla_xu$ & $\jac_g^{-\mathsf T}\nabla_\xi\widehat u$ & $\mathcal G_g[\widehat u]$ \\
Advection & $\bm b\cdot\nabla_xu$ & $(\jac_g^{-1}\widehat{\bm b})\cdot\nabla_\xi\widehat u$ & $\mathcal C_g[\widehat u]$ \\
Diffusion & $\Delta_xu$ & $j_g^{-1}\operatorname{Div}_\xi(\keff_g\nabla_\xi\widehat u)$ & $\mathcal D_g[\widehat u]$ \\
Time derivative & $\partial_tu|_x$ & $\partial_t\widehat u|_\xi-\mathcal A_g[\widehat u]$ & --- \\
Reaction/source & $r(u)+s$ & $r(\widehat u)+s\circ\map_g$ & --- \\
\bottomrule
\end{tabular}
\end{table}

A pointwise reaction retains its functional form, while a prescribed spatial
source is composed with the map. A supplied source field may enter the
candidate library with an unknown multiplier.
For a moving map, Eq.~\eqref{eq:ale-pullback} gives the Eulerian
time-derivative transformation. Writing $\bm w_g=\partial_t\map_g$, the
induced arbitrary Lagrangian--Eulerian (ALE) term is
$\mathcal A_g[\widehat u]=(\jac_g^{-1}\bm w_g)\cdot\nabla_\xi\widehat u$.
It enters with a positive sign on the right-hand side when the
reference-coordinate time derivative is isolated on the left.
If the map follows material motion and the balance uses the corresponding
material-time derivative, this explicit contribution is absorbed.

Each assembled candidate term is assigned one coefficient. Its
coordinate-expanded components are not fitted independently, even when it
contains several reference derivatives and geometry-dependent factors.
State-modulated candidates, such as $\widehat u\mathcal D_g[\widehat u]$,
are formed after assembly and assigned their own coefficients.
In the experiments below, mapped diffusion, ALE transport and their
state-modulated counterparts augment the retained reference-coordinate
library. System-specific libraries and generating balances are detailed in
the equation-discovery protocol below. Hats are suppressed on pulled-back
state fields in those protocols.

\subsection*{Diffusion example of geometry--physics confounding}

\paragraph*{Exact equivalence and scalar attribution.}
The reference domain was the unit square. A scalar deformation level $s$ defined
the orientation-preserving analytic map
\begin{align}
 x(\xi,\eta)&=\xi+s\left[0.11(2\xi-1)\sin(\pi\eta)+0.16\eta\right],\\
 y(\xi,\eta)&=\eta+0.09s(2\eta-1)\sin(\pi\xi).
 \label{eq:diagnostic-map}
\end{align}
The system visualised in Fig.~\ref{fig:confounding}b used a
$101\times81$ grid, $s=1$ and $\kappa=0.01$. A two-peak initial condition with
homogeneous Dirichlet boundaries was advanced to $t=0.55$ by the second-order
Heun method with time step $2\times10^{-4}$. After pull-back, the deformed-domain
and fixed-domain descriptions yield the same discrete equation and therefore
share one integrated trajectory. Their common stored solution gives the exact
zero observation mismatch.

For the fixed-deformation identification experiment,
$s\in\{0,0.1,\ldots,1.2\}$ and nine intrinsic
diffusivities from 0.006 to 0.014 were varied independently. Eight fixed smooth
probe fields were evaluated on the $101\times81$ grid, with boundary-adjacent
points excluded. The misspecified coefficient was the least-squares projection
\begin{equation}
 \widehat\kappa_{\rm Ref}
 =\arg\min_a\sum_m
 \left\|\kappa\mathcal L_g[u_m]-a\Delta_\xi u_m\right\|_2^2,
 \qquad
 \mathcal L_g[u]=j_g^{-1}\nabla_\xi\!\cdot(\keff_g\nabla_\xi u),
 \label{eq:diagnostic-scalar-fit}
\end{equation}
whereas the geometry-complete fit replaced $\Delta_\xi u_m$ by
$\mathcal L_g[u_m]$. Residuals were normalised by the norm of the generated
response. The second pair on the $\widehat\kappa_{\rm Ref}=0.01$ contour was
obtained from the unit-diffusivity projection at $s=1.2$ using the fitted scalar
attribution.

\paragraph*{Five-term operator expansion.}
The prediction experiments used the same map on a $41\times41$ reference grid.
Writing $\keff_g=(K_{ab})_{a,b=1}^{2}$ and expanding
$j_g^{-1}\nabla_\xi\cdot(\keff_g\nabla_\xi u)$ gives
\begin{equation}
 \mathcal L_g[u]=\sum_{i=1}^{5}q_{g,i}D_i[u],
 \quad
 \bm D[u]=
 \begin{bmatrix}u_\xi&u_\eta&u_{\xi\xi}&u_{\xi\eta}&u_{\eta\eta}\end{bmatrix}^{\mathsf T},
 \label{eq:diagnostic-expanded-operator}
\end{equation}
with
\begin{equation}
 \bm q_g=
 \begin{bmatrix}
 j_g^{-1}(\partial_\xi K_{11}+\partial_\eta K_{12})\\
 j_g^{-1}(\partial_\xi K_{12}+\partial_\eta K_{22})\\
 j_g^{-1}K_{11}\\
 2j_g^{-1}K_{12}\\
 j_g^{-1}K_{22}
 \end{bmatrix}.
 \label{eq:diagnostic-q}
\end{equation}
The geometry map and $\keff_g$ were evaluated analytically. Derivatives of the
tensor components in Eq.~\eqref{eq:diagnostic-q} were evaluated by complex-step
differentiation. At $s=0$, $\bm q=(0,0,1,0,1)^{\mathsf T}$, and the expansion
reduces exactly to the reference Laplacian. The prediction target for every
representation was the same noiseless field,
$u_t=\kappa\sum_iq_{g,i}D_i[u]$, with $\kappa=0.01$.

Smooth states were random combinations of 12 low-frequency sine modes with
independent Gaussian coefficients scaled by inverse wavenumber and normalised
to unit root-mean-square amplitude. All five derivatives were evaluated
analytically from the same modal expansion. These exact derivatives provide an
exact-derivative control for state-differentiation error. Prediction was
evaluated using neural representations based on state and coordinates, state
derivatives and geometry-induced coefficient fields. The final representation supplied
the five geometry-induced PDE terms $q_{g,i}D_i[u]$ to a least-squares
calculation. The complete mode set and
implementation details are provided in Section S1.

\paragraph*{Training and evaluation.}
Eight training geometries were uniformly spaced over $s\in[0,1]$, and prediction
was evaluated at the held-out geometry $s=0.75$. The pointwise experiment used
balanced, nested training sets ranging from 800 to 819,200 samples, whereas the
field experiment used balanced, nested training sets ranging from 16 to 8,192
complete fields. Active inputs and targets were standardised using only the
selected training samples at each training size.

For each task and sample count, the three neural representations began from the
same weights and saw the same shuffled samples. The learning curves in
Fig.~\ref{fig:confounding} report the mean and standard deviation across five
runs. Training schedules, exact seeds and the analytical reconstruction check
are reported in Section S1.

%% file: sections/methods/04_02_operator_learning.tex
\subsection*{Operator-learning benchmark design and evaluation}

\paragraph*{Benchmark design and model comparisons.}
We compared baseline and geometry-conditioned models on the Laplace and
nonlinear reaction--diffusion problems from DIMON and the pipe-flow, airfoil
and elasticity benchmarks distributed with Geo-FNO and evaluated with
Transolver \cite{yin2024dimon,li2023geofno,wu2024transolver}.
DIMON and DIMON-Geo used a multi-input branch--trunk formulation, with the
latter supplying geometry-induced fields through dedicated branches.
Transolver-Geo retained the case-specific attention backbone and replaced
physical-coordinate inputs with reference coordinates and geometric channels.
The field choices are motivated by the weak forms above. All models were
trained on solution observations.

Within each baseline--Geo pair, we matched model architectures and training
settings as closely as possible, allowing for the different input
representations and their associated input-processing and architectural
adaptations. Both variants used the same labelled training and test samples,
prediction targets, optimisation schedule and number of epochs or outer
optimisation steps. Training sets comprised the first $n$ samples of the
case-specific training pool and were nested across training sizes, with five
paired random-seed values at each size. Data
partitions, network configurations, and training procedures, etc.,
are specified separately for each case in Supplementary Sections S2--S6.

\paragraph*{Geometry preprocessing.}
Geometric features were constructed from corresponding reference and physical
mesh points. Laplace and reaction--diffusion used the supplied DIMON reference
coordinates and domain correspondences. For pipe flow, airfoil flow and
elasticity, the fixed reference mesh was the pointwise mean of the full
1,000-case training-pool meshes. This reference was reused at every labelled
training size, without using test geometries or additional solution labels.

For each domain correspondence, writing $\bm d_i=x_i-\xi_i$, we estimated the
local displacement gradient by ridge-regularised least squares and recovered
the domain-map Jacobian as
\begin{equation}
 \bm A_i=\arg\min_{\bm A}
 \sum_{j\in\mathcal{N}(i)}
 \left\|(\bm d_j-\bm d_i)-\bm A(\xi_j-\xi_i)\right\|_2^2
 +\lambda\|\bm A\|_F^2,
 \qquad \jac_i=\bm I+\bm A_i.
 \label{eq:jacobian-lsq}
\end{equation}
Here $\mathcal N(i)$ is a neighbourhood in the reference mesh and $\lambda$
is the ridge parameter. The reconstructed Jacobians supplied the local
geometric information for the feature channels. Neighbourhood definitions,
regularisation, mapping-quality diagnostics and case-retention rules are
given in Sections S2--S6. Where feature or target standardisation was used,
its statistics were estimated from the selected training samples and
applied unchanged to the test data. This fitting subset is distinct from
the full geometry pool used to construct the Transolver reference meshes.

\paragraph*{Data-efficiency analysis.}
Training size $n$ counts complete geometry--solution samples. 
For each trained model, the unweighted discrete relative
$L_2$ error was computed for each test sample and then averaged over the
test set. At each measured size $n_i$, the five run-level averages were
summarised by their mean $\overline E_i$ and sample standard deviation $s_i$.
The following procedure was applied to both the DIMON and Transolver
comparisons in Figs.~\ref{fig:operator-dimon}c,e and
\ref{fig:operator-transolver}b,d,f.

Required sample counts were estimated by inverting each model's mean learning
curve using piecewise log--log interpolation. For a target error $E_*$ between
the mean errors at two adjacent measured sizes $n_i$ and $n_{i+1}$,
\begin{equation}
 n(E_*)=\exp\!\left[(1-\theta)\log n_i+\theta\log n_{i+1}\right],
 \qquad
 \theta=\frac{\log E_*-\log\overline E_i}
 {\log\overline E_{i+1}-\log\overline E_i}.
 \label{eq:sample-requirement}
\end{equation}
Only target errors within the common measured error range of the two models
were compared, without extrapolation. The percentage reduction in required
labelled samples was
\begin{equation}
 100\left[1-\frac{n_{\rm Geo}(E_*)}{n_{\rm base}(E_*)}\right],
 \label{eq:sample-saving}
\end{equation}
where the baseline is DIMON or Transolver, as appropriate. These estimates
invert the across-seed mean curves, rather than averaging separately inverted
seed-level curves. They describe interpolated labelled-sample requirements,
not additional training experiments or savings in geometric information or
computational cost.

For visualisation, the mean error was interpolated in log--log space and the
sample standard deviation was interpolated linearly against $\log n$, retaining
its original error units. Shaded bands show the interpolated mean plus or
minus this standard deviation. They represent between-run variation and its
visual interpolation, not confidence intervals or uncertainty estimates for
the inferred sample requirements.

%% file: sections/methods/04_03_discovery_and_evaluation.tex
\subsection*{Geometry-complete equation discovery}

\paragraph*{Identification setting and candidate libraries.}
Equation discovery was evaluated on a deforming thermoelastic solid and a
nutrient-limited tissue undergoing free-boundary growth. In both systems, the
state $u(\xi,t)$ and evolving geometry $x=\map_g(\xi,t)$ were observed, the
geometry-evolution closure was known and the prescribed inputs were collected
in $s^{\rm known}$, excluding all candidate coefficients to be inferred.
Known input fields, such as a source profile, can enter the library with unknown
multipliers. The transport PDE was identified in the reference-coordinate
regression
\begin{equation}
 u_t=\Theta\,\bm c,
\end{equation}
where $\Theta$ denotes the candidate-library matrix and $\bm c$ contains one
coefficient for every column, with zero entries for inactive candidates. A single
vector is shared across all training realizations. These are coefficients of the
chosen equation representation, not a physical-parameter vector. Individual
entries may coincide with physical parameters only for particular candidate
definitions and scalings.

The geometry-complete library retained the reference-coordinate terms and added
the mapped diffusion and ALE transport operators in
Table~\ref{tab:geometry-induced-catalogue}. Both systems used the candidate block
$[\mathcal D_g[u],\,u\mathcal D_g[u],\,\mathcal A_g[u],\,u\mathcal A_g[u]]$.
Here, ``geometry-complete'' refers to this prespecified transformed-operator
block. Both libraries used the same aggregation, normalisation and
model-selection protocol.

\paragraph*{Thermoelastic solid system.}
On a reference annular domain $0.5\le r\le1$, $T(\xi,t)$ denotes temperature, $q(\xi,t)$ the pulled-back heat source, and $\bm d(\xi,t)$ the mechanical displacement field defining $x=\xi+\bm d(\xi,t)$. The known quasi-static elastostatic closure was governed by
\begin{equation}
 \mu_s\Delta\bm d+(\lambda_s+\mu_s)\nabla(\nabla\!\cdot\bm d)
 =\beta_T\nabla T,
\end{equation}
where $\lambda_s$ and $\mu_s$ are Lam{\'e} elastic moduli and $\beta_T$ governs isotropic thermal expansion. The transported temperature obeyed $T_t=\kappa\mathcal D_d[T]+q$ with intrinsic thermal diffusivity $\kappa=0.01$ and unit source gain. The reference grid comprised $25$ radial and $72$ angular points, with $81$ temporal snapshots recorded over $0\le t\le2$.
The geometry-induced candidate block was
$[\mathcal D_d[T],\,T\mathcal D_d[T],\,\mathcal A_d[T],\,T\mathcal A_d[T]]$, where
$\mathcal A_d[T]=(\jac_d^{-1}d_t)\cdot\nabla_\xi T$ was evaluated from the observed displacement velocity. The temperature balance uses the material-time derivative. Consequently, the two ALE candidates have zero generating coefficients. Downstream displacement fields, Jacobian distributions and von Mises stresses were evaluated via the elastostatic closure as validation targets.

\paragraph*{Free-boundary tumour-growth system.}
Let $c(\rho,\theta,t)$ denote biochemical nutrient concentration, $R(\theta,t)$ the local boundary radius, $S$ the continuous nutrient supply, and $\epsilon$ the metabolic uptake field. The physical domain was parameterised in polar coordinates by $x=\rho R(\theta,t)\bm e_r$. On this common coordinate template, nutrient transport was governed by
\begin{equation}
 c_t=D\,\mathcal D_R[c]+\rho\frac{R_t}{R}c_\rho+S-\lambda\epsilon c,
\end{equation}
with intrinsic diffusivity $D=0.012$, unit coefficients for boundary-motion ALE transport and nutrient supply, and metabolic consumption rate $\lambda=0.72$. The moving boundary was generated by the known kinematic closure
\begin{equation}
 R_t=\mu\frac{\epsilon_bc_b}{1+\alpha c_b}-\delta m_b,
\end{equation}
where subscript $b$ indicates boundary values, capturing nutrient-driven mitotic expansion against boundary cell loss. The computational mesh utilized $29$ radial and $96$ angular points, storing $101$ snapshots over $0\le t\le2.4$.
The geometry-induced candidate block was
$[\mathcal D_R[c],\,c\mathcal D_R[c],\,\mathcal A_R[c],\,c\mathcal A_R[c]]$.
Evolving boundary radius, total domain area, and nutrient profiles served as downstream closed-loop validation targets.

\paragraph*{Trajectory partitioning and sparse regression protocol.}
For each benchmark, four evolving-domain realizations generated under distinct forcing configurations and initial geometries were used for identification, while four geometrically shifted realizations were held out for predictive validation. Temporal derivatives and candidate fields were locally aggregated before sparse regression, following a formulation inspired by weak-form methods \cite{messenger2021wsindy}.

Candidate library columns were normalised to unit Euclidean norm. Sparse selection used sequential threshold ridge regression (STRidge), with ridge and threshold parameters selected by leave-one-trajectory-out validation, followed by an ordinary least-squares refit on the selected support \cite{brunton2016sindy,rudy2017pdefind}. Candidate lists, aggregation settings and hyperparameter grids are provided in Section S7.

Noise sensitivity was evaluated at $0\%$, $0.5\%$, $1\%$, $2\%$ and $5\%$
state-field noise across five realizations per level. The trajectory-diversity
analysis evaluated every non-empty subset of the four training trajectories using
the main-experiment hyperparameters. Smoothing and subset protocols are given in
Section S8.

\subsection*{Evaluation and statistical reporting}

For a predicted field $\widehat{y}$ and target $y$ evaluated at the same points,
relative error is
\begin{equation}
 E_{L_2}=\frac{\|\widehat{y}-y\|_2}{\|y\|_2}.
\end{equation}
Operator-learning errors are computed per test sample and then averaged over
the corresponding test set. In the pointwise diffusion experiment, the relative norm
is evaluated across the eight held-out states, whereas in the field task it is averaged
over 16 held-out state fields. Parameter-sweep residuals are computed at each
geometry--diffusivity pair, and PDE-identification prediction errors are
averaged over four held-out trajectories. Stress, displacement, boundary and
area are generated by coupling the identified state equation to the known
geometry closure during closed-loop validation.

For equation residuals, $y$ denotes the sampled time derivative and
$\Theta\widehat{\bm c}$ the derivative reconstructed from the fitted candidate
coefficients, giving $\|\Theta\widehat{\bm c}-y\|_2/\|y\|_2$. Fold reductions
are calculated from unrounded results. Discovery support and coefficient errors
are evaluated on the ordered union of terms in the geometry-complete candidate
library. In this evaluation, $\bm c^\star$ and $\widehat{\bm c}$ denote the
generating and fitted candidate-coefficient vectors on that full ordered union,
with zero entries for inactive terms. An unselected or unavailable candidate has
fitted coefficient zero. A generating term absent from the reference-coordinate
library therefore counts as a false negative. Support $F_1$ is the harmonic mean
of precision and recall, and coefficient relative error is
$\|\widehat{\bm c}-\bm c^\star\|_2/\|\bm c^\star\|_2$.

The exact-equivalence, analytical-reconstruction and clean equation-identification
calculations are deterministic. Diffusion-example neural curves and operator-learning
benchmarks report the mean and standard deviation across five runs. Noise-study
lines show medians across five realizations per level, with individual
realizations displayed as pale points. Trajectory-diversity plots enumerate
every subset.

%% file: sections/06_supplementary.tex
\begingroup
\centering
{\LARGE\bfseries Supplementary Information\par}
\vspace{0.8em}
{\Large\bfseries Geometry--physics confounding impairs PDE learning across varying domains\par}
\vspace{1.0em}
{\normalsize Yinghao Cheng$^{1,2}$, Gengxiang Chen$^{1,3}$, Xu Liu$^{4}$, Qinglu Meng$^{1}$, Yixin Jing$^{1}$, Xiangguo Tang$^{1}$, Wenping Mou$^{1}$, Lihui Wang$^{5,6}$, and Yingguang Li$^{1,*}$\par}
\vspace{0.6em}
{\small
$^{1}$College of Mechanical \& Electrical Engineering, Nanjing University of Aeronautics and Astronautics, Nanjing, China\\
$^{2}$School of Electronics, Electrical Engineering and Computer Science, Queen's University Belfast, Belfast, United Kingdom\\
$^{3}$School of Mechanical and Aerospace Engineering, Queen's University Belfast, Belfast, United Kingdom\\
$^{4}$School of Mechanical and Power Engineering, Nanjing Tech University, Nanjing, China\\
$^{5}$Department of Industrial and Systems Engineering, The Hong Kong Polytechnic University, Hong Kong, China\\
$^{6}$Department of Production Engineering, KTH Royal Institute of Technology, Stockholm, Sweden\\
$^{*}$Correspondence: liyingguang@nuaa.edu.cn.\par
}
\vspace{2.0em}
\endgroup

\setcounter{equation}{0}
\renewcommand{\theequation}{S\arabic{equation}}
\renewcommand{\theHequation}{supp.\arabic{equation}}
\setcounter{figure}{0}
\renewcommand{\theHfigure}{supp.\arabic{figure}}
\setcounter{table}{0}
\renewcommand{\thetable}{S\arabic{table}}
\renewcommand{\theHtable}{supp.\arabic{table}}

\section*{S1. Diffusion implementation details}

The prediction experiments in Fig.~1d compare neural inputs comprising state
and coordinates, exact state derivatives, or state derivatives together with
the five geometry-induced coefficient fields $q_i$ defined in Methods. The
final representation supplies the five geometry-induced PDE terms
$q_iD_i[u]$ directly.

The state-derivative representation uses exact first- and
second-order state derivatives to remove differentiation error while the
geometry-induced coefficient fields remain implicit. The first three representations are matched within
each task. Pointwise inputs are padded to 37 dimensions and use the same
37--64--64--1 GELU MLP. Field inputs are padded to
13 channels and use the same $13\to32\to32\to1$ three-layer $3\times3$ GELU
CNN. Padded channels remain exactly zero. At each count the three neural representations
share initialization, shuffled examples, optimizer, learning-rate schedule,
epoch budget and target. Input and target normalisation is recomputed from the
selected training samples at each training size and then applied to the test data. The final
representation uses a double-precision, five-column least-squares
problem with five independently fitted weights.

Prediction uses a $41\times41$ reference grid. Twelve sine products with mode
pairs from $(1,1)$ to $(4,4)$ generate smooth states. Gaussian modal weights
are attenuated by $(m^2+n^2)^{-1/2}$, and every
state is normalised to unit root-mean-square magnitude. The five derivatives in
the expanded operator are differentiated analytically from the modal series.
The $q_i$ are evaluated from the analytical map, with complex-step
derivatives for the first derivatives of $\bm K$. These analytical derivatives
remove finite-difference error from the comparison between the state-derivative
and geometry-induced-coefficient-field representations.

Eight training geometries are uniformly spaced over $s\in[0,1]$, and prediction
is evaluated at the held-out geometry $s=0.75$.

The pointwise pool contains 100 states, eight geometries and 1,024 nested
interior points per state--geometry pair, or 819,200 samples. The same point
ordering is used for a state on all eight geometries. Nested training sets have sizes
\begin{equation}
 \begin{split}
 \{&800,1{,}600,3{,}200,6{,}400,12{,}800,25{,}600,\\
   &51{,}200,102{,}400,204{,}800,409{,}600,819{,}200\},
 \end{split}
\end{equation}
with each set taking the same number of leading points from all 800
state--geometry blocks. Evaluation uses all $39\times39$ eligible centres from
eight disjoint held-out states.

The field pool contains 1,024 states paired with all eight geometries, or 8,192
complete fields. Nested training sets comprise the first
$16,32,64,128,256,512,1,024,2,048,4,096$ and $8,192$ fields in the stored order.
Because fields are stored in state-major groups of eight, every training set
is exactly balanced across geometry. Sixteen independent states are held
out. The CNN sees the full grid, while training loss and relative error are
computed over the central $35\times35$ region to avoid convolution-padding
effects at the boundary.

All three neural representations are trained for 1,000 epochs with Adam and cosine learning-rate
decay from $10^{-3}$ to $10^{-5}$. Maximum batch sizes are 2,048 point samples
and 32 fields. The extended count curves and error bands report the mean and standard
deviation across five independent optimisation runs.

\section*{Shared settings and reporting conventions\\for operator-learning benchmarks}
\label{sec:si-operator-shared}

Sections S2--S6 specify the data, network architectures and optimisation
protocols for the five operator-learning comparisons. Each training size uses five
runs with seeds $0,7,42,1234,2026$, paired between the baseline and its
geometry-induced counterpart. A training size denotes the number of complete
geometry--solution samples, not the number of spatial query points. Errors are
computed as the unweighted discrete relative $L_2$ norm for each test sample,
averaged over test samples and then summarised by the mean and sample standard
deviation across seeds. Network-width lists below include input and output
dimensions.

\section*{S2. Laplace benchmark}

\paragraph*{Data and geometric representation.}
The Laplace problem uses the DIMON dataset of harmonic solutions on deformed
planar domains with varying Dirichlet boundary data \cite{yin2024dimon}.
Each solution is represented at 2,601 corresponding points on a
$51\times51$ reference grid. The original ordered partition contains 3,303
training cases and 197 test cases. Local Jacobians are reconstructed from the
reference-to-physical displacements using 25 nearest reference points,
including the centre, and ridge parameter $10^{-8}$ in
Eq.~\eqref{eq:jacobian-lsq}. Retaining cases with $\min_\xi j_g>0.5$ within
each partition gives 3,233 training cases and 193 test cases. Their original
order is preserved. Both models use the first
$n\in\{300,500,1000,2000,3000\}$ retained training cases and the same 193
test cases. The reference coordinates, physical meshes, boundary values and
solution targets in the two processed datasets are identical.

The baseline geometry input contains 20 principal-component coefficients,
with ten modes for each Cartesian component of the mapping displacement.
The two principal-component bases and their centring means are fitted anew
on the selected training samples and applied unchanged to the test set.
DIMON-Geo instead receives the three independent symmetric geometry-tensor
channels stored in the processed dataset. Each channel is sampled every
fifth point along both reference-grid axes, giving an $11\times11$ array
and 121 values per branch. This subsampling applies only to the geometric
branch inputs. Both networks predict at all 2,601 query points and receive
the same 68 sampled boundary values. Apart from the baseline PCA centring,
no input or target standardisation is applied.

\paragraph*{Network comparison.}
Both models use fully connected branch--trunk networks with a latent width
of 100. A hyperbolic-tangent activation follows every affine layer in each
branch and trunk, including the final layer. The baseline multiplies the
geometry and boundary branch outputs elementwise and takes their inner
product with the coordinate-trunk output. DIMON-Geo multiplies the outputs
of its three tensor branches and boundary branch before the same trunk
contraction. No additional scalar output bias is used. Table~\ref{tab:si-laplace}
lists the dimensions. The boundary branch and trunk are unchanged, while the
geometry branches differ in both input representation and internal widths.

\begin{table}[htbp]
\centering
\small
\caption{\textbf{Laplace network configurations.} Each tensor component in
DIMON-Geo has a separate branch with the listed dimensions.}
\label{tab:si-laplace}
\renewcommand{\arraystretch}{1.0}
\begin{tabular}{@{}>{\raggedright\arraybackslash}p{0.30\linewidth}>{\raggedright\arraybackslash}p{0.32\linewidth}>{\raggedright\arraybackslash}p{0.30\linewidth}@{}}
\toprule
Component & DIMON & DIMON-Geo \\
\midrule
Geometry branch & $[20,100,100,100]$ & Three $[121,32,32,100]$ branches \\
Boundary branch & $[68,150,150,150,100]$ & $[68,150,150,150,100]$ \\
Tensor-branch fusion & Not used & Elementwise product \\
Coordinate trunk & $[2,100,100,100]$ & $[2,100,100,100]$ \\
\bottomrule
\end{tabular}
\end{table}

\paragraph*{Training and evaluation.}
Both networks are trained using full-batch mean-squared error over the
selected training samples and all spatial points. Training comprises
10,000 Adam steps at a learning rate of $10^{-3}$ and 39,000 steps at
$10^{-4}$, with Adam reinitialised between these stages, followed by 1,000
L-BFGS outer steps. The final trained weights are used for evaluation, with
relative $L_2$ errors computed over all 2,601 spatial points.

\section*{S3. Reaction--diffusion benchmark}

\paragraph*{Data and geometric representation.}
The nonlinear reaction--diffusion benchmark uses the DIMON data on affinely
deformed domains, with varying initial conditions and the no-flux boundary
condition described in Methods \cite{yin2024dimon}. Each sample contains
2,800 spatial points and 26 solution snapshots at
$t=0,0.08,\ldots,2.0$, giving 72,800 space--time queries. The first 2,700 cases
form the training pool and the last 300 form the fixed test set. Training
uses the first $n\in\{500,1000,1500,2000,2700\}$ cases in the pool, giving
nested training sets. The
geometry preprocessing uses 25 nearest reference points, including the
centre, with ridge parameter $10^{-8}$. Its retention criterion is
$\min_\xi j_g>0.4$. The processed dataset contains 3,000 cases and has a
minimum determinant of approximately 0.428.

The baseline reads the supplied 50 initial-condition coefficients and 20
geometry coefficients. These precomputed encodings are not refitted when
the selected training samples change. DIMON-Geo uses 255 sampled initial-condition
values instead of the 50 coefficients, together with four full-resolution
geometric inputs: the determinant field $j_g$ and the three independent
symmetric geometry-tensor channels in the processed dataset. Each geometric
branch therefore receives 2,800 values. Both trunks query the same three
coordinates $(t,\xi_1,\xi_2)$. No input or target standardisation is applied.

\paragraph*{Network comparison.}
The branch and trunk dimensions are listed in
Table~\ref{tab:si-reaction-diffusion}. Each subnetwork uses a hyperbolic-tangent
activation after every affine layer, including its final layer. In the
baseline, the initial-condition and geometry branch outputs are multiplied
elementwise and contracted with the 200-component trunk output.
In DIMON-Geo, the three tensor-branch outputs are concatenated into a
600-component vector and passed through an affine $600\to200$ fusion layer
without a further activation. The fused vector is multiplied elementwise
by the determinant-branch and initial-condition-branch outputs before
contraction with the trunk. The comparison thus changes the initial-condition
encoding and branch dimensions as well as the geometric representation.

\begin{table}[htbp]
\centering
\small
\caption{\textbf{Reaction--diffusion network configurations.}}
\label{tab:si-reaction-diffusion}
\renewcommand{\arraystretch}{1.0}
\begin{tabular}{@{}>{\raggedright\arraybackslash}p{0.30\linewidth}>{\raggedright\arraybackslash}p{0.32\linewidth}>{\raggedright\arraybackslash}p{0.30\linewidth}@{}}
\toprule
Component & DIMON & DIMON-Geo \\
\midrule
Initial-condition branch & $[50,300,300,200]$ & $[255,200,200,200]$ \\
Geometry branch & $[20,200,200,200]$ & Four $[2800,200,200,200]$ branches \\
Tensor-branch fusion & Not used & Affine $600\to200$ \\
Space--time trunk & $[3,200,200,200]$ & $[3,200,200,200]$ \\
\bottomrule
\end{tabular}
\end{table}

\paragraph*{Training and evaluation.}
Both networks are trained using full-batch mean-squared error over the
selected training samples and all spatial points and times. Training comprises
100,000 Adam steps with learning rates $2\times10^{-4}$ for the first
50,000 steps, $10^{-4}$ for the next 40,000 and $10^{-5}$ for the final
10,000, with Adam reinitialised at each transition. The final trained weights
are used for evaluation, with each sample's relative $L_2$ error computed
jointly over all 26 snapshots and 2,800 spatial points.

\section*{S4. Pipe-flow benchmark}

\paragraph*{Data and geometric representation.}
The pipe-flow comparison uses the geometry-varying flow benchmark distributed
with Geo-FNO and Transolver \cite{li2023geofno,wu2024transolver}. The raw
dataset contains 2,310 cases on $129\times129$ grids. We use the first 1,200
cases in source order, with the first 1,000 forming the training pool and
the next 200 forming the test set. Sampling every point along the first grid
axis and every seventh point along the second gives a $129\times19$ grid
with 2,451 points. The scalar target is horizontal velocity $v_x$.
Both models use identical sampled target arrays.

The reference mesh is the pointwise mean of the 1,000 training-pool meshes.
It remains fixed for all training sizes, so small-data runs use geometry
from the full training pool when constructing this reference, but no
additional solution labels or test geometries. Local Jacobians are recovered
using radius-2 structured index patches, containing at most 25 points
including the centre, with truncated patches at the boundary and ridge
parameter $10^{-8}$. No cases are filtered or sorted by Jacobian quality.
All 1,200 selected cases have $j_g>0.2$, with a global minimum of 0.785 and
maximum Jacobian condition number below 4.120.

The baseline receives the two sample-dependent physical coordinates.
Transolver-Geo receives the two reference coordinates and seven additional
channels in the order
$(K_{g,11},K_{g,12},K_{g,22},B_{g,11},B_{g,12},B_{g,21},B_{g,22})$.
The matrices $\keff_g$ and $B_g$ are defined in Methods. Coordinates and
geometric channels are standardised channelwise using means and standard
deviations over the selected training cases and their mesh points. A scalar
target normaliser is fitted over the same training cases. Test data use
these training statistics, with $10^{-8}$ added to each standard deviation.

\paragraph*{Network comparison.}
Both models use the structured-mesh Transolver backbone in
Table~\ref{tab:si-pipe}. The input projection is a two-layer MLP with widths
$[d_{\rm in},256,128]$ and an intermediate GELU activation. Each of the eight
blocks contains pre-normalised physics attention and a pre-normalised
$[128,128,128]$ GELU MLP, both with residual connections. Physics attention
uses eight heads, 64 slice tokens and two $3\times3$ feature projections on
the structured grid. A final layer normalisation and affine $128\to1$ map
produce the scalar prediction. Dropout is zero. Neither time embeddings nor
the optional distance-to-reference-grid positional encoding are used.
The coordinate-only implementation adds a learned 128-component vector
after the input projection. The feature-conditioned structured-mesh path
does not add this vector. The attention backbone is otherwise unchanged.

\begin{table}[htbp]
\centering
\small
\caption{\textbf{Pipe-flow network configurations.}}
\label{tab:si-pipe}
\renewcommand{\arraystretch}{1.0}
\begin{tabular}{@{}>{\raggedright\arraybackslash}p{0.30\linewidth}>{\raggedright\arraybackslash}p{0.32\linewidth}>{\raggedright\arraybackslash}p{0.30\linewidth}@{}}
\toprule
Component & Transolver & Transolver-Geo \\
\midrule
Input channels & 2 physical coordinates & 2 reference coordinates and 7 geometry channels \\
Input MLP & $[2,256,128]$ & $[9,256,128]$ \\
Attention blocks & 8 & 8 \\
Hidden width / heads & 128 / 8 & 128 / 8 \\
Slice tokens / MLP ratio & 64 / 1 & 64 / 1 \\
Output channels & 1 & 1 \\
\bottomrule
\end{tabular}
\end{table}

\paragraph*{Training and evaluation.}
Both models use the first $n\in\{200,400,600,800,1000\}$ cases of the fixed
training pool. Training uses 500 epochs of AdamW with shuffled batches of
four, weight decay $10^{-5}$, gradient-norm clipping at 0.1 and a batch-wise
OneCycleLR schedule with maximum learning rate $10^{-3}$. The loss sums
per-sample relative $L_2$ errors over each batch. Losses and reported errors
are evaluated on the physical velocity scale over all 2,451 spatial points, 
and the final trained weights are used for evaluation.

\section*{S5. Airfoil-flow benchmark}

\paragraph*{Data and geometric representation.}
The airfoil comparison uses the compressible-flow benchmark distributed with
Geo-FNO and Transolver \cite{li2023geofno,wu2024transolver}. Although the raw
dataset contains 2,490 cases, the experiments use only the first 1,200 in
source order. The first 1,000 form the training pool and the following 200
form the fixed test set. Each case retains all $221\times51=11{,}271$ mesh
points, without spatial subsampling. The scalar prediction target is Mach
number, taken from channel 4, using zero-based indexing, of the flow array.

The reference mesh is the pointwise mean of all 1,000 training-pool meshes
and is fixed across training sizes, as in S4. Jacobian reconstruction uses
radius-2 structured patches with ridge parameter $10^{-8}$. The source-order
split is unchanged by the Jacobian diagnostics. No case filtering, sorting
or determinant clipping is applied. The minimum determinant over the 1,200
selected cases is approximately 0.423. The baseline receives physical
coordinates, whereas Transolver-Geo receives reference coordinates and
the four cofactor channels $(C_{g,11},C_{g,12},C_{g,21},C_{g,22})$.
Coordinate, geometry-channel and target normalisation use only the selected
training samples and follow the procedure in S4.

\paragraph*{Network comparison.}
Both models use the same structured-mesh attention blocks, activation,
normalisation, output head and positional-encoding settings as the pipe
models. Their grid dimensions are $221\times51$. The geometry-conditioned
input projection has six input channels rather than the baseline's two.
The learned-vector convention is also the same as in S4. The resulting
configurations are given in Table~\ref{tab:si-airfoil}.

\begin{table}[htbp]
\centering
\small
\caption{\textbf{Airfoil-flow network configurations.} Both models use the
structured-mesh backbone specified in S4.}
\label{tab:si-airfoil}
\renewcommand{\arraystretch}{1.0}
\begin{tabular}{@{}>{\raggedright\arraybackslash}p{0.30\linewidth}>{\raggedright\arraybackslash}p{0.32\linewidth}>{\raggedright\arraybackslash}p{0.30\linewidth}@{}}
\toprule
Component & Transolver & Transolver-Geo \\
\midrule
Input channels & 2 physical coordinates & 2 reference coordinates and 4 cofactor channels \\
Input MLP & $[2,256,128]$ & $[6,256,128]$ \\
Attention blocks & 8 & 8 \\
Hidden width / heads & 128 / 8 & 128 / 8 \\
Slice tokens / MLP ratio & 64 / 1 & 64 / 1 \\
Output channels & 1 & 1 \\
\bottomrule
\end{tabular}
\end{table}

\paragraph*{Training and evaluation.}
Both models use the first $n\in\{200,400,600,800,1000\}$ cases of the fixed
training pool. Training uses 500 epochs, shuffled batches of four and the
AdamW, gradient-clipping and batch-wise OneCycleLR settings specified in S4. The loss sums
per-sample relative $L_2$ errors over each batch. Losses and reported errors
are computed from decoded Mach numbers over all 11,271 spatial points. 
The final trained weights are used for evaluation.

\section*{S6. Elasticity benchmark}

\paragraph*{Data and geometric representation.}
The elasticity comparison uses the varying-unit-cell benchmark distributed
with Geo-FNO and Transolver \cite{li2023geofno,wu2024transolver}. The source
dataset contains 2,000 cases, each represented by a 972-point unstructured
mesh. The first 1,000 cases form the training pool, and the final 200 cases,
with zero-based source indices 1,800--1,999, form the test set. The intervening
800 cases are unused. The target is the scalar stress field, and both models 
predict this same field at all 972 points.

The fixed reference point cloud is the pointwise mean of the 1,000
training-pool meshes. Jacobians are reconstructed from 24 nearest reference
points, including the centre, with ridge parameter $10^{-8}$. The reference
mesh and neighbour stencils do not depend on the test geometries and are
reused at every training size. The 1,200 selected cases are retained in their
prescribed order without Jacobian-based filtering. Their minimum determinant
is approximately 0.405, so all exceed 0.2. This value is a diagnostic of the
selected dataset, not a criterion used to change its membership.

The baseline receives unnormalised physical coordinates. Transolver-Geo
receives the two reference coordinates and six channels ordered as
$(C_{g,11},C_{g,12},C_{g,21},C_{g,22},d_{g,1},d_{g,2})$, where
$\bm d_g=x-\xi$ describes the benchmark domain correspondence, not the
predicted mechanical response. The reference coordinates and six geometric
channels are standardised using the selected training samples. Both models
use a scalar stress normaliser fitted to those samples.

\paragraph*{Network comparison.}
Both models use the irregular-mesh Transolver variant, with eight blocks,
hidden width 128, eight heads, 64 slice tokens, MLP ratio one, GELU
activation and zero dropout. The block structure and scalar output head
follow S4, but the two feature projections in physics attention are pointwise
affine maps rather than structured-grid convolutions. No time or optional
distance-based positional encoding is used. In this irregular-mesh
implementation, both input paths add the learned 128-component vector after
their input MLP. The attention backbone is held fixed and the input widths
are listed in Table~\ref{tab:si-elasticity}.

\begin{table}[htbp]
\centering
\small
\caption{\textbf{Elasticity network configurations.} Both models use
irregular-mesh physics attention.}
\label{tab:si-elasticity}
\renewcommand{\arraystretch}{1.0}
\begin{tabular}{@{}>{\raggedright\arraybackslash}p{0.30\linewidth}>{\raggedright\arraybackslash}p{0.32\linewidth}>{\raggedright\arraybackslash}p{0.30\linewidth}@{}}
\toprule
Component & Transolver & Transolver-Geo \\
\midrule
Input channels & 2 physical coordinates & 2 reference coordinates and 6 geometry channels \\
Input MLP & $[2,256,128]$ & $[8,256,128]$ \\
Attention blocks & 8 & 8 \\
Hidden width / heads & 128 / 8 & 128 / 8 \\
Slice tokens / MLP ratio & 64 / 1 & 64 / 1 \\
Output channels & 1 & 1 \\
\bottomrule
\end{tabular}
\end{table}

\paragraph*{Training and evaluation.}
Both models use the first $n\in\{200,400,600,800,1000\}$ cases of the fixed
training pool. Training uses 500 epochs, shuffled batches of one and the
AdamW, gradient-clipping and batch-wise OneCycleLR settings specified in S4. The relative $L_2$
training loss and reported errors are evaluated on the physical stress scale
over all 972 spatial points. The final trained weights are used for evaluation.

\section*{S7. Equation-discovery protocols and identified equations}

\paragraph*{Shared identification protocol.}
The thermoelastic and tumour-growth cases use the same sparse-identification
protocol. For each system, four trajectories are used for identification and
four geometrically shifted trajectories are held out for predictive validation.
The transported-state equation is identified while the geometry-evolution
closure remains known.

Both systems use a five-point fourth-order centred finite difference for time
derivatives. Candidate columns and target derivatives are averaged over local
$3\times5$ spatial boxes before equidistant subsampling, and candidate columns
are normalised to unit Euclidean norm. Sequential threshold ridge regression
(STRidge) \cite{brunton2016sindy,rudy2017pdefind} is evaluated over ridge penalties
$\{10^{-10},10^{-8},10^{-6},10^{-5}\}$ and relative thresholds
$\{0.002,0.005,0.01,0.02,0.04,0.06,0.08,0.14\}$. Leave-one-trajectory-out
validation selects the most parsimonious model whose mean residual is within
$20\%$ of the minimum, with ties resolved by residual magnitude. The selected
support is then refitted by ordinary least squares across all four training
trajectories.

Support and coefficient errors follow the definitions in Methods.
Coefficients are indexed over the full evaluation library, with zeros for
inactive terms, and refer to the chosen candidate representation rather than a
physical-parameter vector. In both cases, the geometry-complete library retains
the reference-coordinate candidates and adds mapped diffusion, ALE transport
and their state-modulated counterparts. The case-specific libraries and fitted
equations are given below.

\paragraph*{Thermoelastic system.}
Identification targets the temperature law in Fig.~\ref{fig:thermal-discovery},
with the mechanical response supplied by the known elastic closure. The
prescribed heat-source field $q$ belongs to $s^{\rm known}$, whereas its
multiplier is an unknown candidate coefficient.

The reference-coordinate candidate library is
\begin{equation*}
 [1,T,T^2,T^3,q,T_{\xi_1},T_{\xi_2},\Delta_\xi T,T\Delta_\xi T,|\nabla_\xi T|^2].
\end{equation*}
The geometry-complete candidate library adds
\begin{equation*}
 [\mathcal D_d[T],T\mathcal D_d[T],\mathcal A_d[T],T\mathcal A_d[T]],
\end{equation*}
where $\mathcal A_d[T]=(\jac_d^{-1}d_t)\cdot\nabla_\xi T$ is evaluated from the
observed displacement velocity. The material-time
formulation assigns zero generating coefficients to the two thermal ALE
candidates, which remain available to the sparse selector.

The fitted reference-coordinate equation was
\begin{equation}
 T_t=0.999061q+0.014685\Delta_\xi T-0.049062\,T\Delta_\xi T,
\end{equation}
and the geometry-complete equation was
\begin{equation}
 T_t=0.999980q+0.00999978\,\mathcal D_d[T].
\end{equation}
The reference-coordinate and geometry-complete fits have support $F_1$ scores
of 0.4 and 1 and held-out residuals of 0.171 and $3.02\times10^{-5}$,
respectively. The geometry-complete coefficient relative $L_2$ error is
$2.03\times10^{-5}$. Both fitted temperature equations were integrated with
the same elastic closure. Fig.~\ref{fig:thermal-discovery}
and Extended Data Fig.~3 report the resulting held-out predictions.

\paragraph*{Tumour-growth system.}

The nutrient law in Fig.~\ref{fig:tumour-discovery} is identified while the boundary-growth closure
remains known. The reference-coordinate candidate library is

\begin{equation*}
 [1,c,c^2,c^3,S,-\epsilon c,c_\rho,\rho^{-1}c_\theta,
 \Delta_{\rm ref}c,c\Delta_{\rm ref}c,|\nabla_{\rm ref}c|^2].
\end{equation*}
The geometry-complete candidate library adds
\begin{equation*}
 [\mathcal D_R[c],c\mathcal D_R[c],\mathcal A_R[c],c\mathcal A_R[c]].
\end{equation*}
The prescribed fields $S$ and $\epsilon$ belong to $s^{\rm known}$, while the
multipliers of $S$ and $-\epsilon c$ are candidate coefficients to be inferred.
These terms remain in the physical block because they acquire no new
differential geometry field in the pointwise strong form.

The fitted reference-coordinate equation was
\begin{equation}
 c_t=0.109157c^2-0.128395c^3+0.987610S+0.732135(-\epsilon c)
      +0.062263c_\rho+0.008024\Delta_{\rm ref}c,
\end{equation}
and the geometry-complete equation was
\begin{equation}
 c_t=1.000001S+0.719997(-\epsilon c)+0.0120013\,\mathcal D_R[c]
     +1.000067\,\mathcal A_R[c].
\end{equation}
The reference-coordinate and geometry-complete fits have support $F_1$ scores
of 0.4 and 1, coefficient relative $L_2$ errors of 0.640 and
$4.25\times10^{-5}$, and held-out residuals of 0.180 and
$6.93\times10^{-4}$, respectively. Both fitted nutrient equations were
integrated with the same boundary-growth closure.
Fig.~\ref{fig:tumour-discovery} and Extended Data Fig.~4 report the resulting held-out predictions.

\clearpage
\section*{S8. Sensitivity to noise and training-set size\\in equation discovery cases}

Gaussian noise at 0, 0.5\%, 1\%, 2\% and 5\% of the state-field standard deviation is applied with five seeds per level. Geometry remains noise-free. Non-zero-noise fields are smoothed sequentially in time, radius and angle with third-order Savitzky--Golay filters using windows 7, 5 and 7 before locally aggregated STRidge \cite{messenger2021wsindy}.

Median thermoelastic support F1 remains 1 through 5\% noise, and neither inactive thermal ALE candidate is selected in any realization. Tumour support remains exact through 1\% and decreases at 2\% and 5\%. A separate clean-data analysis enumerates all subsets of one to four training realizations using fixed main-experiment hyperparameters. One trajectory recovers exact support in both evolving-domain systems.

Extended Data Fig.~5 reports the three identification metrics at each noise
level. Thermoelastic reference-library residuals remained about an order of
magnitude larger throughout. Tumour reference residuals were larger through
2\% noise. At 5\%, both libraries selected the same degraded sparse model and
had the same median residual. Extended Data Fig.~6 reports all clean-data
training subsets.

\renewcommand{\figurename}{Extended Data Fig.}

\begin{figure}[p]
\centering
\includegraphics[width=\linewidth]{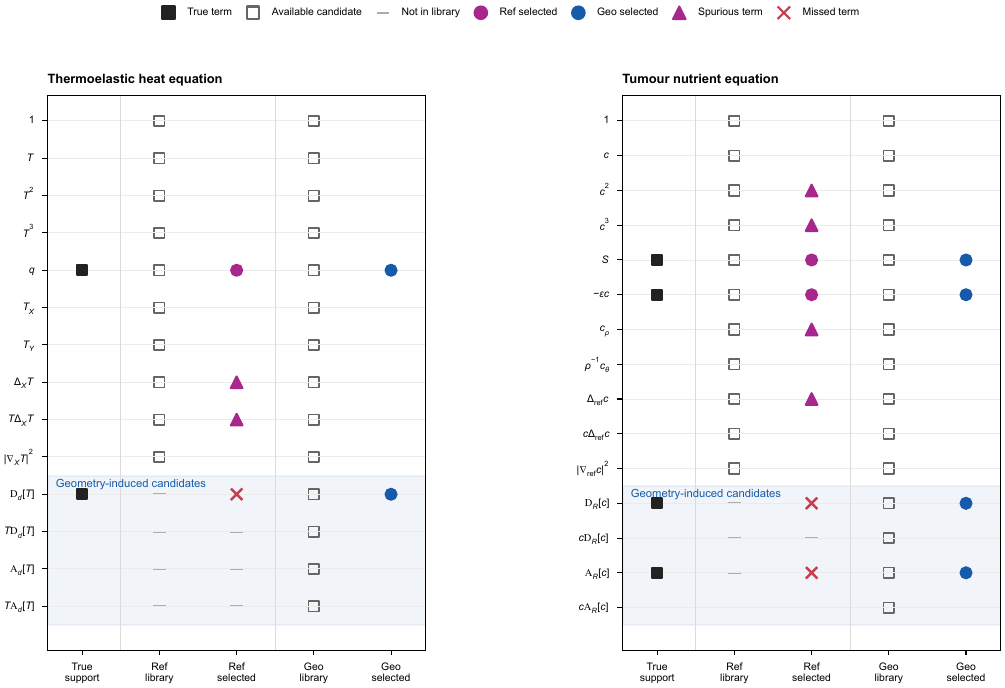}
\caption{\textbf{Candidate-library availability and selection.} Term-by-term comparison of candidate construction and sparse selection. The geometry-complete library for each system contains the same four-term template: mapped diffusion, state-modulated mapped diffusion, ALE transport and state-modulated ALE transport. For each system, the first column marks the generating support. Ref and Geo denote the reference-coordinate and geometry-complete libraries, each shown through library-membership and selected-support columns. Open squares mark available candidates and grey dashes mark terms absent from a library. Purple and blue circles mark terms selected by Ref and Geo, respectively. Purple triangles mark additional Ref selections, and red crosses mark missed generating terms.}
\label{edfig:library-audit}
\end{figure}
\clearpage

\begin{figure}[p]
\centering
\includegraphics[width=\linewidth]{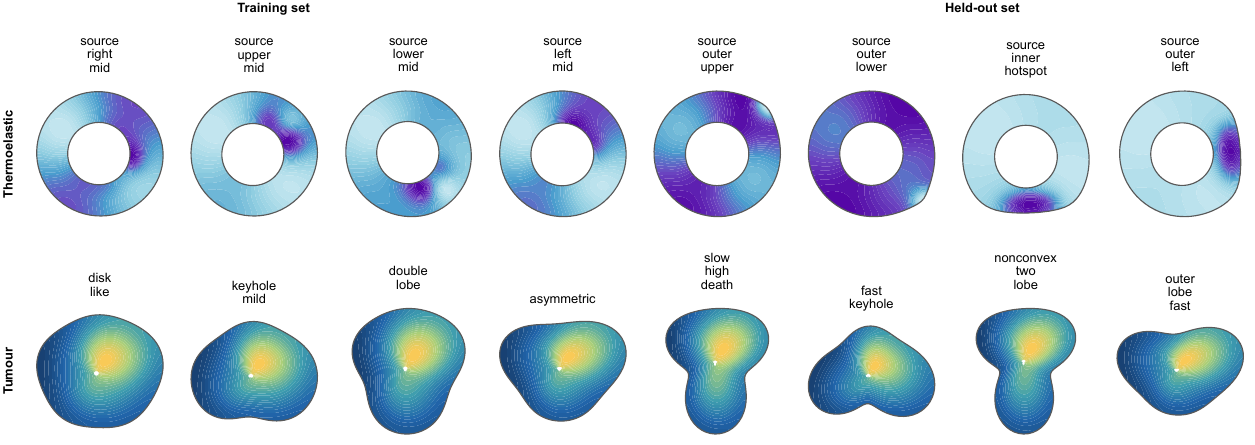}
\caption{\textbf{Training and held-out evolving-domain realizations.} Final-state gallery for the four training and four held-out trajectories in each system. The held-out set changes source placement and geometric response beyond the combinations used for sparse identification.}
\label{edfig:geometry-gallery}
\end{figure}
\clearpage

\begin{figure}[p]
\centering
\includegraphics[width=\linewidth]{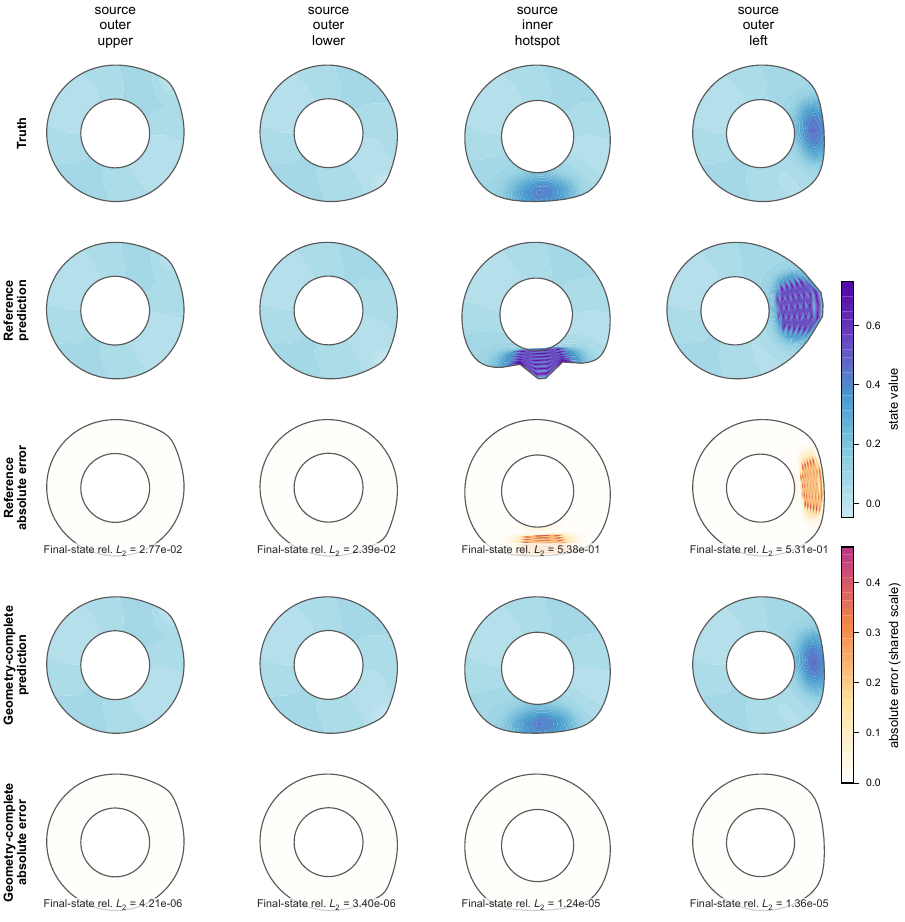}
\caption{\textbf{Thermoelastic transfer across all held-out realizations.} Final-state temperature truth, reference-coordinate prediction and absolute error, and geometry-complete prediction and absolute error for all four held-out trajectories. Prediction rows share a physical-value colour scale and error rows share an absolute-error scale. Labels report final-state relative $L_2$ error. The known elastic closure is used in both integrations.}
\label{edfig:thermal-transfer}
\end{figure}

\begin{figure}[p]
\centering
\includegraphics[width=\linewidth]{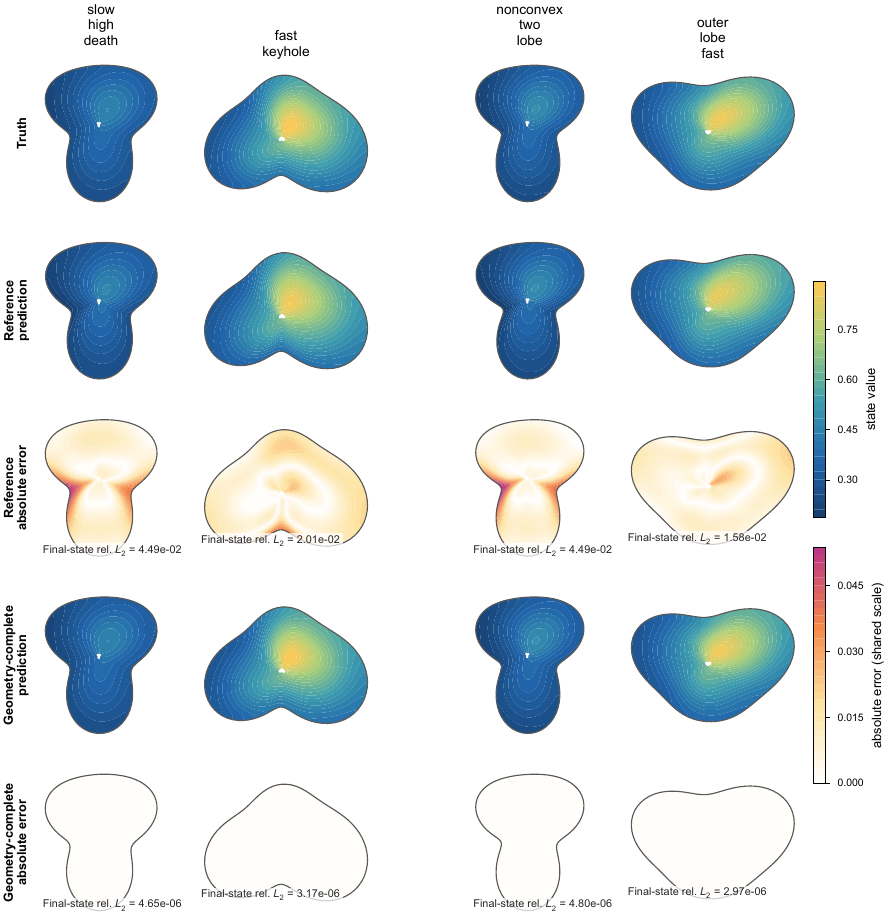}
\caption{\textbf{Tumour transfer across all held-out realizations.} Final-state nutrient truth, reference-coordinate prediction and absolute error, and geometry-complete prediction and absolute error for all four held-out trajectories. Prediction rows share a physical-value colour scale and error rows share an absolute-error scale. Labels report final-state relative $L_2$ error. The known moving-boundary closure is used in both integrations.}
\label{edfig:tumor-transfer}
\end{figure}

\begin{figure}[p]
\centering
\includegraphics[width=\linewidth]{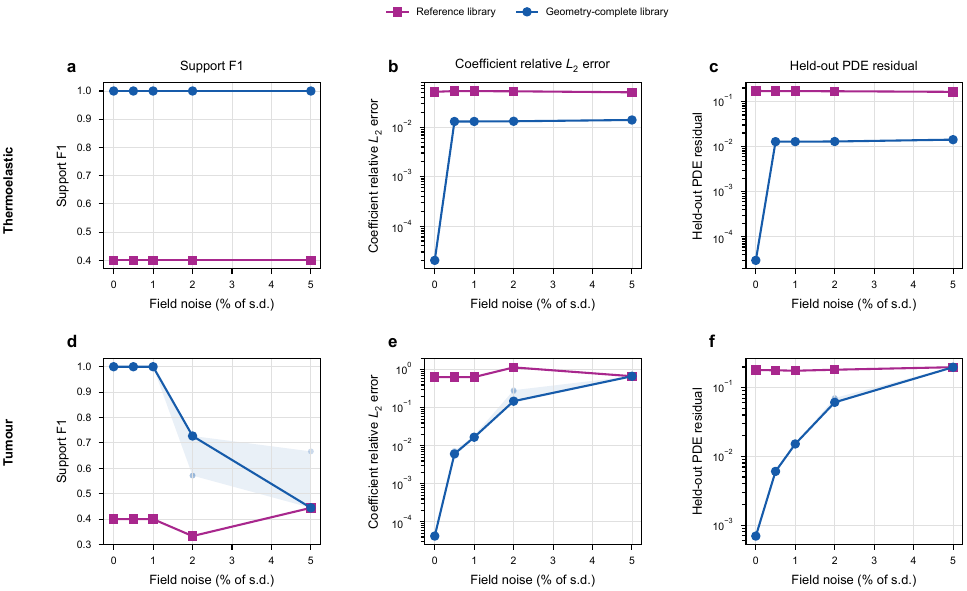}
\caption{\textbf{Sensitivity to state-field noise.} Support $F_1$, coefficient relative $L_2$ error and clean held-out PDE residual after identification from state fields with 0--5\% additive noise relative to the field standard deviation. Large symbols show medians across five noise realizations and pale points show individual realizations. Geometry variables remain noise-free.}
\label{edfig:noise-robustness}
\end{figure}
\clearpage

\begin{figure}[p]
\centering
\includegraphics[width=\linewidth]{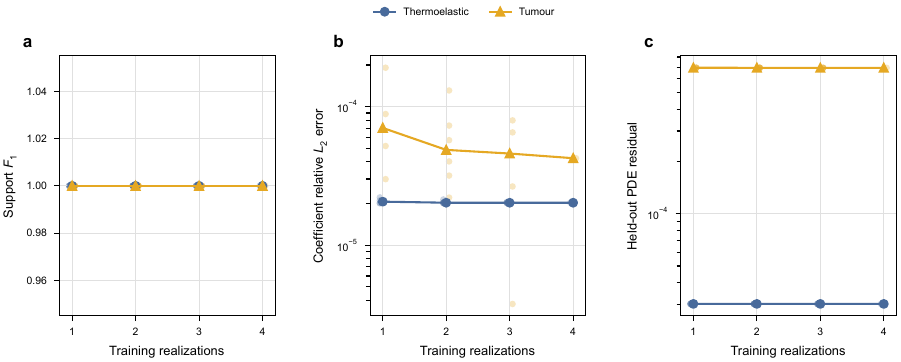}
\caption{\textbf{Training-set size and subset stability.} Support $F_1$, coefficient relative $L_2$ error and held-out PDE residual versus the number of clean training realizations for the geometry-complete candidate library. All subsets of one to four trajectories are enumerated. Large symbols show the subset median and pale points show individual subsets.}
\label{edfig:training-diversity}
\end{figure}
\clearpage